\documentclass[lettersize,journal]{IEEEtran}
\usepackage{amsmath,amsfonts}
\usepackage{array}
\usepackage{textcomp}
\usepackage{stfloats}
\usepackage{url}
\usepackage{verbatim}
\usepackage{graphicx}
\usepackage{cite}

\newtheorem{myDef}{Definition}
\usepackage{xcolor}
\usepackage{multirow}
\usepackage[most]{tcolorbox}
\usepackage{subfig}
\usepackage{booktabs}
\usepackage{threeparttable}

\usepackage[linesnumbered,ruled,vlined]{algorithm2e}

\begin{document}

\title{Towards Irreversible Machine Unlearning for Diffusion Models}

\author{
Xun Yuan,
        Zilong Zhao,
        Jiayu Li,
     Aryan Pasikhani,\\
     Prosanta Gope,~\IEEEmembership{Senior Member,~IEEE}
        and~Biplab~Sikdar,~\IEEEmembership{IEEE Fellow}
\thanks{Xun Yuan and Biplab Sikdar are with the Department of Electrical and Computer Engineering, College of Design and Engineering, National University of Singapore, Singapore. (E-mail: e0919068@u.nus.edu, bsikdar@nus.edu.sg).}
\thanks{Zilong Zhao and Jiayu Li are with Betterdata, Singapore. (E-mail: zilong@betterdata.ai, jiayu@betterdata.ai).}
\thanks{Aryan Pasikhani and Prosanta Gope are with the Department of Computer Science, University of Sheffield, United Kingdom. (E-mail: aryan.pasikhani@sheffield.ac.uk, p.gope@sheffield.ac.uk).}
}

\markboth{
Preprint}%
{Shell \MakeLowercase{\textit{et al.}}: A Sample Article Using IEEEtran.cls for IEEE Journals}


\maketitle

\begin{abstract}
Diffusion models are renowned for their state-of-the-art performance in generating synthetic images. However, concerns related to safety, privacy, and copyright highlight the need for machine unlearning, which can make diffusion models forget specific training data and prevent the generation of sensitive or unwanted content. Current machine unlearning methods for diffusion models are primarily designed for conditional diffusion models and focus on unlearning specific data classes or features. Among these methods, finetuning-based machine unlearning methods are recognized for their efficiency and effectiveness, which update the parameters of pre-trained diffusion models by minimizing carefully designed loss functions. However, in this paper, we propose a novel attack named Diffusion Model Relearning Attack (DiMRA), which can reverse the finetuning-based machine unlearning methods, posing a significant vulnerability of this kind of technique. Without prior knowledge of the unlearning elements (a class of objects or features), DiMRA optimizes the unlearned diffusion model on an auxiliary dataset to reverse the unlearning, enabling the model to regenerate previously unlearned elements. To mitigate this vulnerability, we propose a novel machine unlearning method for diffusion models, termed as Diffusion Model Unlearning by Memorization (DiMUM). Unlike traditional methods that focus on forgetting, DiMUM memorizes alternative data or features to replace targeted unlearning data or features in order to prevent generating such elements.
In our experiments, we demonstrate the effectiveness of DiMRA in reversing state-of-the-art finetuning-based machine unlearning methods for diffusion models, highlighting the need for more robust solutions. We extensively evaluate DiMUM, demonstrating its superior ability to preserve the generative performance of diffusion models while enhancing robustness against DiMRA.
\end{abstract}


\section{Introduction}
Diffusion models (DMs) have emerged as a powerful class of generative models, attracting significant attention for their ability to produce high-quality synthetic images. DMs are broadly categorized into two types: unconditional and conditional. Unconditional DMs generate data that follows the distribution of the training data without any additional input other than Gaussian noise \cite{ho2020denoising,nichol2021improved}. In contrast, conditional DMs (CDMs) generate content based on specific input conditions, such as text descriptions \cite{saharia2022photorealistic} or reference images \cite{lugmayr2022repaint,saharia2022palette}, enabling controlled, directed content generation. Due to their impressive generative capabilities and versatility, CDMs are applied in various domains, including content creation \cite{saharia2022photorealistic}, image editing \cite{lugmayr2022repaint,saharia2022palette}, and medical imaging \cite{kazerouni2023diffusion,wolleb2022diffusion}.

Despite the remarkable generative capabilities of CDMs, their use raises several concerns related to AI-generated content, including the production of harmful outputs \cite{schramowski2023safe}, copyright infringement \cite{mittica2023ai}, and privacy violations \cite{golda2024privacy}. For instance, commercial generative models should be designed to avoid producing images that conflict with local cultural norms or violate copyright laws. Moreover, privacy regulations such as the General Data Protection Regulation (GDPR) \cite{EU_GDPR_2016} grant individuals the right to request the deletion of their personal information from the training data of generative models. Consequently, machine unlearning (MU) techniques have attracted growing research interest. They are being actively explored within the scope of CDMs to unlearn improper training data and thus mitigate these potential societal risks. Existing MU methods for CDMs primarily aim to prevent the generation of images from undesired classes \cite{huang2024unified,heng2024selective,fan2023salun}, unwanted styles \cite{gandikota2024unified,kumari2023ablating,gandikota2023erasing,zhang2024forget}, or with undesirable features \cite{wu2024erasediff,li2024get,wu2024scissorhands,lyu2024one}. Broadly, these MU methods can be categorized into two approaches: filter-based and finetuning-based. Filter-based MU methods for CDMs \cite{gandikota2024unified,lyu2024one,li2024get,zhang2024forget} work by filtering or modifying the conditioning input associated with the unlearning concepts. On the other hand, finetuning-based methods for CDMs work by optimizing the pre-trained CDMs' parameters to increase the training loss of the undesired data \cite{huang2024unified} or replace the undesired concepts with benign concepts \cite{wu2024scissorhands,kumari2023ablating,fan2023salun,gandikota2023erasing,wu2024erasediff}.  

Filter-based MU methods for CDMs have inherent limitations. For example, filters can be easily removed at the code level \cite{li2024safegen}, making them ineffective in open-source models. Besides, as demonstrated in \cite{zhang2024unlearncanvas}, filter-based MU methods for CDMs often suffer from either poor unlearning effectiveness or low-quality generation after unlearning, whereas finetuning-based methods effectively address both challenges simultaneously. Consequently, finetuning-based methods are considered more efficient and effective for MU in CDMs. However, this paper shows that existing finetuning-based MU methods can be reversed even when the attacker lacks prior knowledge of unlearning elements. Specifically, we propose a novel relearning attack named Diffusion Model Relearning Attack (DiMRA). DiMRA reverses the finetuning-based MU methods by optimizing the unlearned CDMs on an auxiliary dataset that contains no data related to the unlearned elements. Experimental results, e.g., Fig. \ref{fig:demo}, demonstrate that DiMRA can reverse existing finetuning-based MU methods for CDMs, enabling the CDM to regenerate previously unlearned elements (harmful or unwanted).

The failure of finetuning-based MU methods stems from two key reasons: (i) their unlearning losses are non-convergent, and (ii) their retain losses constrain the CDM’s parameters to remain close to those of the pre-trained model during unlearning. As a result, the parameters of the unlearned CDM remain close to the pre-trained model’s parameters and fail to reach a local optimum. Further optimization of the unlearned CDM on an auxiliary dataset can reverse its parameters toward the model before unlearning, and thus the reversed CDM generates previously unlearned elements. On the other hand, the memorization and generalization behaviours of DMs have been studied in \cite{baptista2025memorization, yoon2023diffusion, gu2023memorization, zhang2024emergence}. According to the analysis in these papers, the generation capacity of DMs stems from the (partial) memorization of training data.
Building on this insight, we propose Diffusion Model Unlearning by Memorization (DiMUM), a novel MU method for CDMs. Given a conditioning input from the unlearning set, DiMUM optimizes the CDM to memorize alternative images rather than the original images associated with the conditioning input. According to the theoretical analysis in \cite{baptista2025memorization}, the CDM will then memorize these new images, and thus forget the original images after the unlearning process. As a result, the unlearned CDM prevents the regeneration of the unlearning elements. Furthermore, the loss function of DiMUM is convergent, which leads to the parameters of the unlearned CDM reaching a new local optimum, thus enhancing the robustness against DiMRA.

In summary, the paper has the following contributions: 
\begin{itemize} 
    \item We propose a {\emph{novel} relearning attack}, Diffusion Model Relearning Attack (DiMRA), which can reverse the unlearning process of finetuning-based MU methods for CDMs, enabling the CDM to regenerate previously unlearned elements (harmful or unwanted).

    \item We propose a {\emph{novel} MU method for CDMs}, Diffusion Model Unlearning by Memorization (DiMUM), which achieves unlearning by memorizing alternative elements to replace those being unlearned. 

    \item We provide a comprehensive evaluation of the proposed DiMRA and DiMUM. Experimental results show that DiMRA can successfully attack existing finetuning-based MU methods for CDMs. Besides, DiMUM shows superior performance in preserving the generative performance and improving the robustness against DiMRA.
    
\end{itemize}

\section{Related Work}

This section presents a literature review on machine unlearning for diffusion models and attacks for machine unlearning in diffusion models.

\subsection{Machine Unlearning in Diffusion Models}

\subsubsection{Filter-based MU for CDMs}
In text-to-image latent diffusion models, text conditioning is incorporated into the network through text embeddings, which are extracted from pre-trianed embedding model. \cite{gandikota2024unified} modifies the weights of the cross-attention layers to induce specific changes to selected text embeddings corresponding to edited concepts while minimizing the impact on preserved concepts. 
Forget-Me-Not~\cite{zhang2024forget} introduces an attention steering loss, which first computes the attention maps between input features and context embeddings associated with the forgetting concept, then minimizes the attention maps.
\cite{li2024get} accomplishes the same objective in two steps. The \textit{Soft-weighted Regularization} step leverages singular value decomposition to remove undesired concepts from the text embedding. The \textit{Inference-time Text Embedding Optimization} step introduces new loss functions that guide the cross-attention layer to emphasize the desired concepts and suppress the generation of unlearned concepts.
SPM~\cite{lyu2024one} introduces a lightweight, trainable adapter for DM’s neural network that enables efficient concept erasure. Using a latent anchoring method, the authors combine a surrogate and target concept for finetuning, with an objective function that includes erasing and anchoring losses to minimize impact on retained concepts.
It is worth noting that our proposed attack, DiMRA, is not designed to target filter-based MU for CDMs, but rather to target finetuning-based approaches.

\subsubsection{Finetuning-based MU for CDMs}

Another branch of MU for CDMs works by finetuning the pre-trained CDM. For instance, Salun~\cite{fan2023salun} enables unlearning by finetuning only the most affected salient weights, using randomly labelled data from the unlearning set to update these salient weights.
Erased Stable Diffusion (ESD)~\cite{gandikota2023erasing} finetunes the model to align the conditional scores of undesired concepts with those of unconditioned concepts, permanently removing learned concepts. 
\cite{kumari2023ablating} opts to replace the target concept with a user-defined concept, typically a more general term. For instance, it fine-tunes the model to generate "a cat" instead of "a grumpy cat."
In \cite{wu2024erasediff}, the authors modify the reverse process of DMs by updating the loss function to align the predicted noise of specific concepts with a predefined noise distribution.
\cite{wu2024scissorhands} employs gradient-based techniques to map out the weight saliency within UNet related to the concept to be unlearned, concentrating fine-tuning efforts on the most sensitive neurons associated with the target concept.
\cite{huang2024unified} applies the gradient-ascent method and improves the stability during unlearning by embedding the unlearning update within a remain-preserving manifold. While gradient ascent is commonly used in MU for classification models \cite{thudi2022unrolling,10113700} and large language models \cite{yao2024large,cha2025towards}, it is less favoured for diffusion models (DMs) due to its detrimental impact on generative performance.

\subsection{Attacks for Machine Unlearning in Diffusion Models}

Recent studies \cite{chin2023prompting4debugging, maus2023blackboxadversarialprompting, zhang2024generate} compromise MU methods for CDMs using carefully designed adversarial prompts. These studies iteratively optimize the CDM's text prompts until the unlearned CDM generates the targeted elements that should have been unlearned. However, these methods \cite{chin2023prompting4debugging, maus2023blackboxadversarialprompting, zhang2024generate} require prior knowledge of the unlearning concept and its associated images, making them less practical in adversarial settings with limited background knowledge. Furthermore, the attacks proposed in \cite{chin2023prompting4debugging} and \cite{maus2023blackboxadversarialprompting} require an auxiliary CDM and a pre-trained classifier, respectively. In contrast, the proposed DiMRA reverses the MU methods to enable the unlearned CDM to regenerate unlearning elements without the need for prior knowledge or auxiliary models.

\begin{figure*} 
\center{\includegraphics[width=0.85\linewidth, scale=1.]{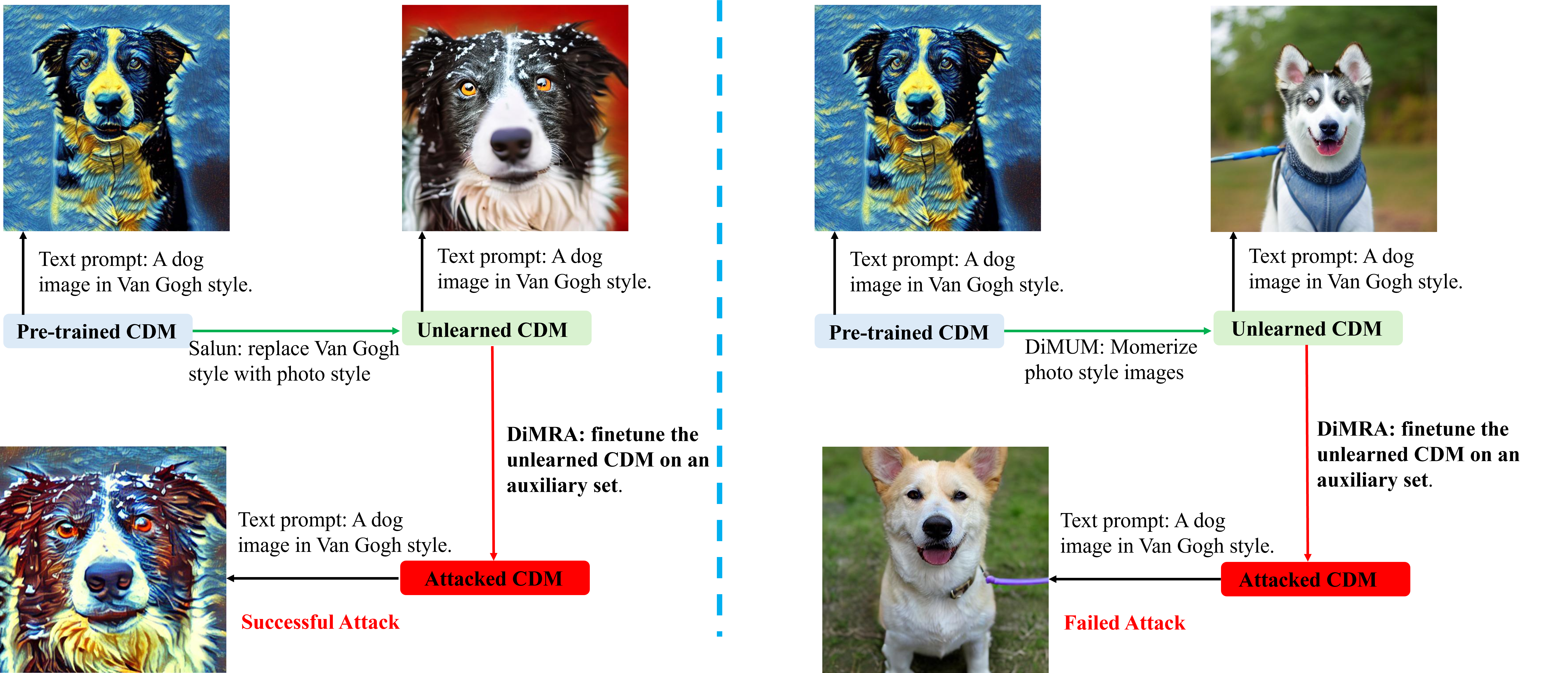}}
\caption{
Workflow of DiMRA. The left panel illustrates the reversal of Salun by the proposed DiMRA when unlearning the Van Gogh style. The right panel shows DiMUM’s resilience under identical conditions.}
\label{fig:demo}
\end{figure*}

\section{Preliminaries}

This section briefly introduces the CDMs and finetuning-based MU methods for CDMs to better understand the proposed DiMRA and DiMUM.

\subsection{Conditional Diffusion Models}

The CDMs introduced in this section are based on DDPM \cite{ho2020denoising} and omit DDIM \cite{song2020denoising} since they have the same loss function.
CDMs operate in two stages, i.e., the forward and reverse processes. The forward process starts from clean data $\boldsymbol{x}_0$ and iteratively adds Gaussian noise to the data for $T$ steps until the data $\boldsymbol x_T$ becomes nearly indistinguishable from pure Gaussian noise. Given a forward process step $t$, $\boldsymbol x_t$ can be calculated by,
\begin{equation}
    \label{eq:forward process}
    \boldsymbol x_t\left(\boldsymbol{x}_0, \boldsymbol{\epsilon}\right)=\sqrt{\bar{\alpha}_t} \boldsymbol{x}_0+\sqrt{1-\bar{\alpha}_t} \boldsymbol{\epsilon},
\end{equation}
where $\bar{\alpha}_t$ is a pre-defined parameter, $\boldsymbol{\epsilon} \sim \mathcal{N}(\boldsymbol{0}, \boldsymbol{I})$, and $t \in \{1,2,\cdots, T\}$.

The reverse process starts from pure Gaussian noise $\hat{\boldsymbol{x}}_T\sim \mathcal{N}(\boldsymbol{0}, \boldsymbol{I})$ and iteratively denoises the data with the estimated noise $\boldsymbol{\hat{\epsilon}}$ calculated by Classifier-Free Guidance (CFG) \cite{ho2022classifier}, 
\begin{equation}
\label{eq:estimated noise}
\boldsymbol{\hat{\epsilon}} = \boldsymbol{\epsilon_\theta}\left(\boldsymbol{\hat{x}}_t, t, \emptyset\right)+\beta \left(\boldsymbol{\epsilon_\theta}\left(\boldsymbol{\hat{x}}_t, t, \boldsymbol{c}\right)-\boldsymbol{\epsilon_\theta}\left(\boldsymbol{\hat{x}}_t, t , \emptyset\right)\right), 
\end{equation}
where $\boldsymbol{\epsilon_{\theta}}$ is a trainable approximator, $\boldsymbol{c}$ is the conditioning feature, $\emptyset$ is the feature for the null condition and usually a zero vector, and $\beta$ is a scalar for the conditional scale. Then, given $\boldsymbol{\hat{x}}_t$, $\boldsymbol{\hat{x}}_{t-1}$ can be calculated by,
\begin{equation}
\label{eq:reverse process}
\boldsymbol{\hat{x}}_{t-1}=\frac{1}{\sqrt{\alpha_t}}\left(\boldsymbol{\hat{x}}_t-\frac{1-\alpha_t}{\sqrt{1-\bar{\alpha}_t}} \boldsymbol{\hat{\epsilon}}\right)+\sigma_t \boldsymbol{z} ,
\end{equation}
where $t \in \{1,2,\cdots, T\}$, $\boldsymbol{z} \sim \mathcal{N}(\boldsymbol{0}, \boldsymbol{I})$, and $\sigma_t$ and $\alpha_t$ are pre-defined parameters.

The approximator $\boldsymbol{\epsilon_{\theta}}$ is optimized by the following loss function,
\begin{equation}
\label{eq:ori_loss_c}
L(\boldsymbol{\epsilon_\theta})=\mathbb{E}_{t, (\boldsymbol{x}_0,\boldsymbol{c})\in D, \boldsymbol\epsilon, \boldsymbol{I})}\left[\left\|\boldsymbol{\epsilon}-\boldsymbol{\epsilon}_{\boldsymbol{\theta}}(\boldsymbol{x}_t|\boldsymbol{c})\right\|^2\right], 
\end{equation}
where $\boldsymbol{x}_0$ is training data, $\boldsymbol{c}$ is conditioning input for $\boldsymbol{x}_0$, $t\in \{1,2,\cdots, T\}$, and $\boldsymbol{\epsilon} \sim \mathcal{N}(\boldsymbol{0}, \boldsymbol{I})$.

\subsection{Machine Unlearning for CDMs}

This section begins by outlining the objective of MU for CDMs. Next, we present the core mechanisms of existing finetuning-based MU methods for CDMs.

\subsubsection{Objective definition:} 
The objective of MU for CDMs is defined in Definition \ref{def:obj}.

\begin{myDef} \label{def:obj}
Let $\boldsymbol{\theta}_p$\footnote{In this paper, $\boldsymbol{\theta}$ denotes all parameters of a CDM, including $\boldsymbol{\epsilon_{\theta}}$, $\alpha_t$, $\bar{\alpha}_t$, and $\sigma_t$, where $t\in \{1,2,\cdots,T\}$.} represent a CDM pre-trained on dataset $D$. Given an unlearning set $D_u$ containing unlearning elements (e.g. feature and class), a retain set $D_r$ without unlearning elements, and an MU mechanism $\mathcal{M}$ for CDM, the unlearned CDM $\boldsymbol{\theta}_u$ is obtained by $\boldsymbol{\theta}_u=\mathcal{M}(\boldsymbol{\theta}_p,D_r,D_u)$. Let $P_u$ denote the distribution of the images generated by $\boldsymbol{\theta}_u$ and $P_{target}$ denote the distribution of the images generated by $\boldsymbol{\theta}_p$ discarding the images with unlearning elements. The objectives of $\mathcal{M}$ are to (i) increase the similarity between $P_u$ and $P_{target}$ and (ii) prevent generating images that contain unlearning elements. 
\end{myDef}

In general, it is assumed that the distribution of $D_r$ is similar to that of $P_{target}$ in Definition \ref{def:obj}. Thus, the similarity between $P_u$ and $P_{D_r}$ is applied in practice to evaluate objective (i) in Definition \ref{def:obj}.  

\subsubsection{Mechanisms of existing MU for CDMs:} 

Finetuning-based MU methods for CDMs first construct an unlearning set $D_u$ and a retain set $D_r$. The unlearning set $D_u$ consists of images related to the unlearning concept or contains the unlearning feature, while the retain set $D_r$ includes benign images that are to be preserved. 


Some methods, such as \cite{huang2024unified}, minimize the loss function in \textbf{Eq.}\eqref{eq:ori_loss_c} when $\boldsymbol{x_0}$ is sampled from $D_r$ to preserve the CDM's generative capacity, while maximizing it when $\boldsymbol{x_0}$ is sampled from $D_u$ to degrade the CDM's ability to generate images resembling those in $D_u$.
This loss function can be calculated by,
\begin{equation}
\begin{aligned}
\label{eq:unlearn_loss_1}
L_{1}(\boldsymbol{\epsilon_\theta})&=\beta\mathbb{E}_{t, (\boldsymbol{x}_0,\boldsymbol{c}_r)\in D_r, \boldsymbol\epsilon}\left[\left\|\boldsymbol{\epsilon}-\boldsymbol{\epsilon}_{\boldsymbol{\theta}}(\boldsymbol{x}_t|\boldsymbol{c}_r)\right\|^2\right] \\
&-\mathbb{E}_{t, (\boldsymbol{x}_0,\boldsymbol{c}_u)\in D_u, \boldsymbol\epsilon}\left[\left\|\boldsymbol{\epsilon}-\boldsymbol{\epsilon}_{\boldsymbol{\theta}}(\boldsymbol{x}_t|\boldsymbol{c}_u)\right\|^2\right],
\end{aligned}
\end{equation}
where $D_u$ is the unlearning set, $D_r$ is the retain set, and $\beta$ is a balancing coefficient. The first term in \textbf{Eq.}\eqref{eq:unlearn_loss_1} is commonly referred to as the `retain loss', while the second term is known as the `unlearning loss'.

Other methods, such as \cite{gandikota2023erasing, wu2024scissorhands, kumari2023ablating,wu2024erasediff,fan2023salun}, also minimize the loss function \textbf{Eq.}\eqref{eq:ori_loss_c} when $\boldsymbol{x}_0\in D_r$ to preserve the CDM's generative capability. Therefore, their retain loss is the same as that in \textbf{Eq.}\eqref{eq:unlearn_loss_1}.
As for the unlearning loss, this kind of method applies the benign feature to replace the unlearning feature. Specifically, given a data pair $(\boldsymbol{x}_u,\boldsymbol{c}_u)$ sampled from the unlearning set, these methods assign an artificial benign condition $\boldsymbol{c}_r^\prime$ to $\boldsymbol{x}_u$ and then apply $\boldsymbol{\epsilon}_{\boldsymbol{\theta}}(\boldsymbol{x}_t|\boldsymbol{c_r^\prime})$ as the label to optimize $\boldsymbol{\epsilon}_{\boldsymbol{\theta}}(\boldsymbol{x}_t|\boldsymbol{c_u})$, where $\boldsymbol{x}_t$ is the noised $\boldsymbol{x_u}$ after $t$ forward steps. The loss function of this kind of methods can be calculated by,
\begin{equation}
\begin{aligned}
\label{eq:unlearn_loss_2}
L_{2}(\boldsymbol{\epsilon_\theta})&=\beta\mathbb{E}_{t, (\boldsymbol{x}_0,\boldsymbol{c}_r)\in D_r, \boldsymbol\epsilon}\left[\left\|\boldsymbol{\epsilon}-\boldsymbol{\epsilon}_{\boldsymbol{\theta}}(\boldsymbol{x}_t|\boldsymbol{c}_r)\right\|^2\right] \\
&+\mathbb{E}_{t, (\boldsymbol{x}_0,\boldsymbol{c}_u)\in D_u, \boldsymbol\epsilon}\left[\left\|\boldsymbol{\epsilon}^s_{\boldsymbol{\theta}}(\boldsymbol{x}_t|\boldsymbol{c_r^\prime})-\boldsymbol{\epsilon}_{\boldsymbol{\theta}}(\boldsymbol{x}_t|\boldsymbol{c}_u)\right\|^2\right],
\end{aligned}
\end{equation}
where superscript $s$ means stopping gradients during optimization. 
\textbf{Note:} Although various methods incorporate distinct technical details, their core mechanisms can be distilled into the loss functions \textbf{Eq.}\eqref{eq:unlearn_loss_1} and \textbf{Eq.}\eqref{eq:unlearn_loss_2}.

\section{Diffusion Model Relearning Attack (DiMRA)}

The overall attack workflow of DiMRA is illustrated in Fig. \ref{fig:demo}. The unlearning feature in this figure is the Van Gogh style. First of all, the original CDM $\boldsymbol{\theta}_p$ is trained on dataset $D$, which includes Van Gogh-style images. Second, we construct an unlearning set $D_u$ that contains only Van Gogh-style images and a retain set $D_r$ that does not contain Van Gogh-style images. Then, the unlearned model is obtained by a MU method $\mathcal{M}$, i.e., $\boldsymbol{\theta}_u=\mathcal{M}(\boldsymbol{\theta}_p,D_r,D_u)$, which prevents the generation of Van Gogh-style images. 
Next, we apply DiMRA to attack the unlearned CDM by finetuning it on an auxiliary dataset. \textbf{Note} that the auxiliary dataset does not contain any image related to the Van Gogh style. If the attack is successful, the attacked CDM will generate Van Gogh-style images given the conditioning feature $\boldsymbol{c}_u$ of the unlearn set $D_u$. The left side of Fig. \ref{fig:demo} demonstrates a successful attack targeting the CDM unlearned by Salun \cite{fan2023salun}, while the right side shows the robustness of the proposed DiMUM against DiMRA.

\subsection{Attacker Model} \label{sec:Attacker Model}

\textbf{Attacker's Capabilities.} 
We assume that the attacker has access to the parameters of the targeted CDM. This assumption is realistic, as model providers often release pre-trained models, which are typically publicly available and may not be sufficiently protected (e.g., Stable Diffusion 1.4\footnote{\url{https://huggingface.co/CompVis/stable-diffusion-v1-4}}). Additionally, we assume that the attacker is familiar with the prompt format, including available classes and textual prompts, which is commonly documented alongside the CDM when released for public use. 

\textbf{Attacker's Background Knowledge.} 
We evaluate the DiMRA under two background knowledge assumptions:

Assumption (i): The attacker has access to the retain set $D_r$. This represents the most favourable scenario for the attacker, where the retain set $D_r$ has been leaked.

Assumption (ii): The attacker has access to an auxiliary dataset that follows a similar distribution to the retain set $D_r$. This assumption is realistic and can be easily achieved, e.g., by collecting the generated synthetic images.

\textbf{Attacker's Goal.} Given a CDM that has undergone proper unlearning, the attacker aims to reverse the unlearning process, i.e., make the CDM regenerate previously forgotten elements, thereby compromising the MU methods and potentially exposing sensitive or unsafe information that should have been erased from the CDMs.

\subsection{Attack Strategy} \label{sec:dimra}

In DiMRA, the attacker first constructs an auxiliary dataset $D_{au}$ based on Assumption (i) or (ii). Then, the attacker update the unlearned CDM's parameters by minimizing the following loss function,
\begin{equation}
\label{eq:ori_dimra}
L_{DiMRA}(\boldsymbol{\epsilon_\theta})=\mathbb{E}_{t, {(\boldsymbol{x}_0,\boldsymbol{c}) \in D_{au}}, \boldsymbol\epsilon}\left[\left\|\boldsymbol{\epsilon}-\boldsymbol{\epsilon}_{\boldsymbol{\theta}}(\boldsymbol{x}_t|\boldsymbol{c})\right\|^2\right],
\end{equation}
which is similar to the loss function of CDMs \textbf{Eq.}\eqref{eq:ori_loss_c} but the training data $(\boldsymbol{x}_0,\boldsymbol{c})$ is sampled from the auxiliary dataset $D_{au}$ instead of the original training dataset $D$. Optimizing the unlearned CDM by minimizing 
$L_{DiMRA}$ for a certain number of steps can cause the CDM to regenerate previously unlearned elements.

\textbf{Rational behind DiMRA.} 
Let's closely investigate the core mechanisms of the existing MU methods for CDMs, i.e., \textbf{Eq.}\eqref{eq:unlearn_loss_1} and \textbf{Eq.}\eqref{eq:unlearn_loss_2}. (i) The unlearning loss (second term) of \textbf{Eq.}\eqref{eq:unlearn_loss_1} aims to maximize the original loss function \textbf{Eq.}\eqref{eq:ori_loss_c} of CDM when input $\boldsymbol{x}_0$ is sampled from the unlearning set, resulting in the unlearning loss in \textbf{Eq.}\eqref{eq:unlearn_loss_1} being divergent. (ii) The unlearning loss (second term) of \textbf{Eq.}\eqref{eq:unlearn_loss_2} changes the optimization objective of $\boldsymbol{\epsilon}_{\boldsymbol{\theta}}(\boldsymbol{x}_t|\boldsymbol{c}_u)$ from the ground-truth $\boldsymbol{\epsilon}$ to $\boldsymbol{\epsilon}^s_{\boldsymbol{\theta}}(\boldsymbol{x}_t|\boldsymbol{c_r^\prime})$ when input $\boldsymbol{x}_0$ is sampled from the unlearning set. This change leads to non-convergence of the unlearning loss. For example, when $t$ is small, the noised image $\boldsymbol{x}_t$ retains much of the information of the original image $\boldsymbol{x}_0$ associated with unlearning feature $\boldsymbol{c}_u$ and thus $\boldsymbol{\epsilon}^s_{\boldsymbol{\theta}}(\boldsymbol{x}_t|\boldsymbol{c_r^\prime})$ is meaningless. 
The divergence and non-convergence of the unlearning losses in \textbf{Eq.}\eqref{eq:unlearn_loss_1} and \textbf{Eq.}\eqref{eq:unlearn_loss_2} are empirically validated the experimental section.

Now, Let's assume that the pre-trained approximator $\boldsymbol{\epsilon}_{\boldsymbol{\theta}_p}$ has reached a local optimum by minimizing the original loss function of CDMs \textbf{Eq.}\eqref{eq:ori_loss_c}. 
The objectives of MU for CDMs are to achieve a new local optimum by minimizing the retain loss of \textbf{Eq.}\eqref{eq:unlearn_loss_1} and \textbf{Eq.}\eqref{eq:unlearn_loss_2} while preventing the generation of unlearning elements.
The unlearning losses in \textbf{Eq.}\eqref{eq:unlearn_loss_1} and \textbf{Eq.}\eqref{eq:unlearn_loss_2} push the approximator's parameters away from the original local optimum, thereby preventing the CDM from generating the unlearning elements. 
Besides, the retain loss in \textbf{Eq.}\eqref{eq:unlearn_loss_1} and \textbf{Eq.}\eqref{eq:unlearn_loss_2} keeps the approximator's parameters close to those of the pre-trained approximator to preserve the generative capability of the CDM.
Unfortunately, the unlearning losses in \textbf{Eq.}\eqref{eq:unlearn_loss_1} and \textbf{Eq.}\eqref{eq:unlearn_loss_2} hinder the approximator's parameters from converging to a new local optimum due to their divergence and non-convergence as discussed above.
As a result, optimizing the approximator, which have been unlearned with \textbf{Eq.}\eqref{eq:unlearn_loss_1} or \textbf{Eq.}\eqref{eq:unlearn_loss_2}, by minimizing the loss function \textbf{Eq.}\eqref{eq:ori_dimra} of DiMRA can thus push the parameters of the approximator back to the original local optimum, thereby relearning the previously unlearned elements.

\section{Diffusion Model Machine Unlearning by Memorization (DiMUM)}

We have analyzed that the existing MU mechanisms for CDMs fail to optimize the CDMs' parameters to a new local optimum w.r.t. the retain loss in \textbf{Eq.}\eqref{eq:unlearn_loss_1} and \textbf{Eq.}\eqref{eq:unlearn_loss_2}. Consequently, DiMRA can reverse the parameters of the unlearned CDMs back to the original local optimum of the pre-trained models by finetuning them according to \textbf{Eq.}\eqref{eq:ori_dimra}. In this section, we introduce the proposed DiMUM that can address this problem.

DiMUM applies the retain loss as \textbf{Eq.}\eqref{eq:unlearn_loss_1} and \textbf{Eq.}\eqref{eq:unlearn_loss_2} in order to achieve objective (i) in Definition \ref{def:obj}, i.e.,
\begin{equation}
\label{eq:retain_loss_dimum}
L_r(\boldsymbol{\epsilon_\theta})=\mathbb{E}_{t, (\boldsymbol{x}_0,\boldsymbol{c})\in D_r, \boldsymbol\epsilon}\left[\left\|\boldsymbol{\epsilon}-\boldsymbol{\epsilon}_{\boldsymbol{\theta}}(\boldsymbol{x}_t|\boldsymbol{c})\right\|^2\right].
\end{equation}
However, the retain loss \textbf{Eq.}\eqref{eq:retain_loss_dimum} does not guarantee the erasure of the information of unlearning set $D_u$ from the pre-trained model. According to the analysis in \cite{baptista2025memorization,yoon2023diffusion,gu2023memorization,zhang2024emergence}, the generation ability of the DMs stems from their memorization or partial memorization of the training data. Based on these findings, we can erase the information of $D_u$ from the pre-trained CDM by simply memorizing alternative elements when providing the CDM with the conditioning features in $D_u$. 
To achieve this, we construct a dataset $D_u^\prime$ by combining the images samples from the retain set $D_r$ and conditioning features of the unlearning set $D_u$, i.e.,
\begin{equation}
\begin{aligned}
\label{eq:construction_new_dataset}
D_u^\prime = \{(\boldsymbol{x}_1,\boldsymbol{c}_1), \cdots (\boldsymbol{x}_N,\boldsymbol{c}_N)\} \quad
\text{where } \boldsymbol{x}_i \in D_r, \boldsymbol{c}_i \in D_u.
\end{aligned}
\end{equation}
Then, the unlearning loss, $L_u(\boldsymbol{\epsilon_\theta})$, of the proposed DiMUM is calculated by,
\begin{equation}
\label{eq:unlearn_loss_dimum}
L_u(\boldsymbol{\epsilon_\theta})=\mathbb{E}_{t,(\boldsymbol{x}_0, \boldsymbol{c})\in D_u^\prime, \boldsymbol\epsilon}\left[\left\|\boldsymbol{\epsilon}-\boldsymbol{\epsilon}_{\boldsymbol{\theta}}(\boldsymbol{x}_t|\boldsymbol{c})\right\|^2\right].
\end{equation}
The unlearning loss of DiMUM \textbf{Eq.}\eqref{eq:unlearn_loss_dimum} is similar to \textbf{Eq.}\eqref{eq:ori_loss_c}, \textbf{Eq.}\eqref{eq:ori_dimra}, and \textbf{Eq.}\eqref{eq:retain_loss_dimum}, whose specificity lies in that the input $(\boldsymbol{x}_0, \boldsymbol{c})$ is sampled from the newly constructed $D_u^\prime$ with \textbf{Eq.}\eqref{eq:construction_new_dataset}. After the convergence of the CDM by minimizing \textbf{Eq.}\eqref{eq:unlearn_loss_dimum}, the unlearned CDM will generate images following the distribution of $D_r$ given the prompts $\boldsymbol{c}$ of the unlearning set $D_u$ and thus prevent generating images that contain unlearning elements.
The loss function of DiMUM combines the retain loss \textbf{Eq.}\eqref{eq:retain_loss_dimum} and unlearning loss \textbf{Eq.}\eqref{eq:unlearn_loss_dimum}, i.e.,
\begin{equation}
\label{eq:loss_dimum}
L_{DiMUM}= \beta L_u(\boldsymbol{\epsilon_\theta}) + L_u(\boldsymbol{\epsilon_\theta}), 
\end{equation}
where $\beta$ is a balancing coefficient. The workflow of the proposed DiMUM is summarized in Algorithm \ref{alg:proposed DiMUM}.

\begin{algorithm}
\caption{Workflow of DiMUM.} \label{alg:proposed DiMUM}
\SetKwInOut{Input}{Input}
\SetKwInOut{Output}{Output}
\Input{Retain set $D_r$; Unlearning set $D_u$; pre-trained approximator $\boldsymbol{\epsilon_\theta}$; $\boldsymbol{\theta}_{\epsilon_\theta}$ denotes parameters of $\boldsymbol{\epsilon_\theta}$}
{\footnotesize
\Output{Unlearned approximator $\boldsymbol{\epsilon_\theta}$}

$D_u^\prime = \{(\boldsymbol{x}_1,\boldsymbol{c}_1), \cdots, (\boldsymbol{x}_N,\boldsymbol{c}_N)\}
\text{ , } \boldsymbol{x}_i \in D_r, \boldsymbol{c}_i \in D_u.$

\For{$n \in \{1,2,\cdots,N_\text{unlearn}\}$}  
{
     \tcp*[h]{\textit{{Compute retain loss \textbf{Eq.}\eqref{eq:retain_loss_dimum}}}}\;

     Sample $(\boldsymbol{x}_0, \boldsymbol{c})$ from $D_r$

     Sample $\boldsymbol{\epsilon}$ from Gaussian distribution $\mathcal{N}(\mathbf{0}, \mathbf{I})$
     
    $L_r=\left\|\boldsymbol{\epsilon}-\boldsymbol{\epsilon}_{\boldsymbol{\theta}}\left(\sqrt{\bar{\alpha}_t} \boldsymbol{x}_0+\sqrt{1-\bar{\alpha}_t} \boldsymbol{\epsilon}, t, \boldsymbol{c}\right)\right\|^2$ 

    \tcp*[h]{\textit{{Compute unlearning loss \textbf{Eq.}\eqref{eq:unlearn_loss_dimum}}}}\;

    Sample $(\boldsymbol{x}_0, \boldsymbol{c})$ from $D_u^\prime$

     Sample $\boldsymbol{\epsilon}$ from Gaussian distribution $\mathcal{N}(\mathbf{0}, \mathbf{I})$
     
    $L_u=\left\|\boldsymbol{\epsilon}-\boldsymbol{\epsilon}_{\boldsymbol{\theta}}\left(\sqrt{\bar{\alpha}_t} \boldsymbol{x}_0+\sqrt{1-\bar{\alpha}_t} \boldsymbol{\epsilon}, t, \boldsymbol{c}\right)\right\|^2$ 

    \tcp*[h]{\textit{{Compute the whole loss \textbf{Eq.}\eqref{eq:loss_dimum}}}}\;

    $L_{DiMUM}= \beta L_u(\boldsymbol{\epsilon_\theta}) + L_{r}(\boldsymbol{\epsilon_\theta})$

    $\boldsymbol{\theta}_{\epsilon_\theta} \leftarrow \boldsymbol{\theta}_{\epsilon_\theta} - \eta \nabla_{\boldsymbol{\theta}_{\epsilon_\theta}} L_{DiMUM}$ 
}
}
\textbf{Return} $\boldsymbol{\epsilon_\theta}$

\end{algorithm}

\textbf{Core difference between the proposed DiMUM and existing MU methods for CDMs.} As discussed above, DiMRA leverages the non-convergence of loss functions in existing MU methods to reverse their effects. Additionally, such non-convergence can impair the generative performance of the unlearned CDMs.
In contrast, both the unlearning and retain losses of DiMUM follow the same structure as the original loss function \textbf{Eq.}\eqref{eq:ori_loss_c} of CDM, which has been shown to be convergent in \cite{ho2020denoising}. Consequently, after unlearning with DiMUM, the CDM converges to a new approximated local optimum w.r.t. the retain loss (\textbf{Eq.}\eqref{eq:retain_loss_dimum}), which also serves as a local optimum for the DiMRA loss function (\textbf{Eq.}\eqref{eq:ori_dimra}).
As a result, it becomes significantly more difficult for DiMRA to compromise a CDM that has been unlearned using DiMUM. Here we assume the unlearning loss \textbf{Eq.}\eqref{eq:unlearn_loss_dimum} has limited impact on the convergence of the retain loss since the images in $D_u^\prime$ follows a similar distribution to those in $D_r$.

\section{Experiments} \label{sec:experiments}
In this section, we first evaluate the performance of the proposed DiMRA and DiMUM on small-scale CDMs trained on the CIFAR-10 dataset, where a specific class is unlearned, the same as \cite{fan2023salun, huang2024unified}. We then assess their performance on large-scale CDMs using the benchmark introduced in \cite{zhang2024unlearncanvas}, where an art style is unlearned.

\begin{figure*} 
  \centering
  \subfloat[Airplane]
  {\includegraphics[width=0.2\linewidth]{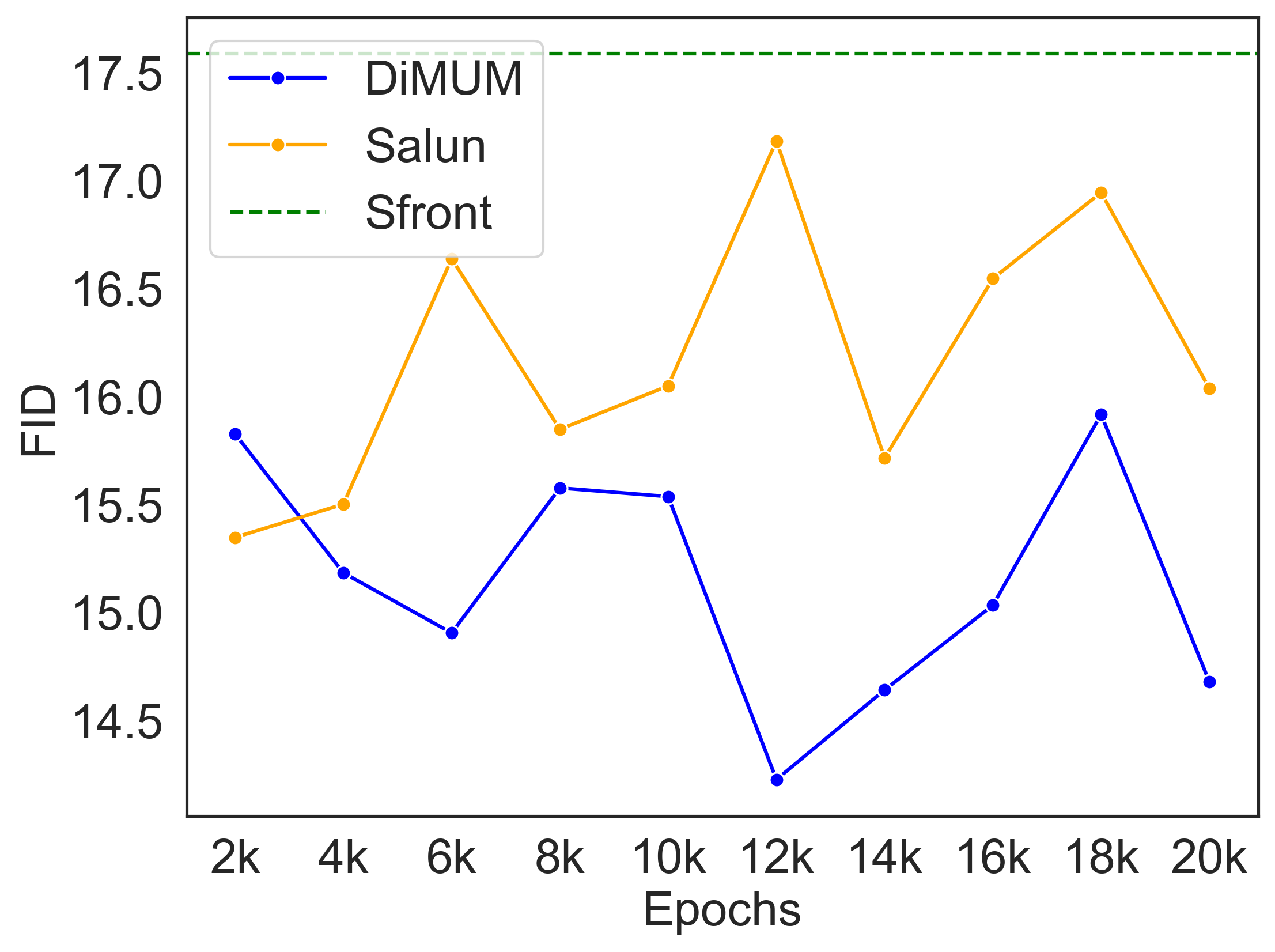}\label{fig:Unlearn_Class_0_FID}}     
  \subfloat[Automobile]
  {\includegraphics[width=0.2\linewidth]{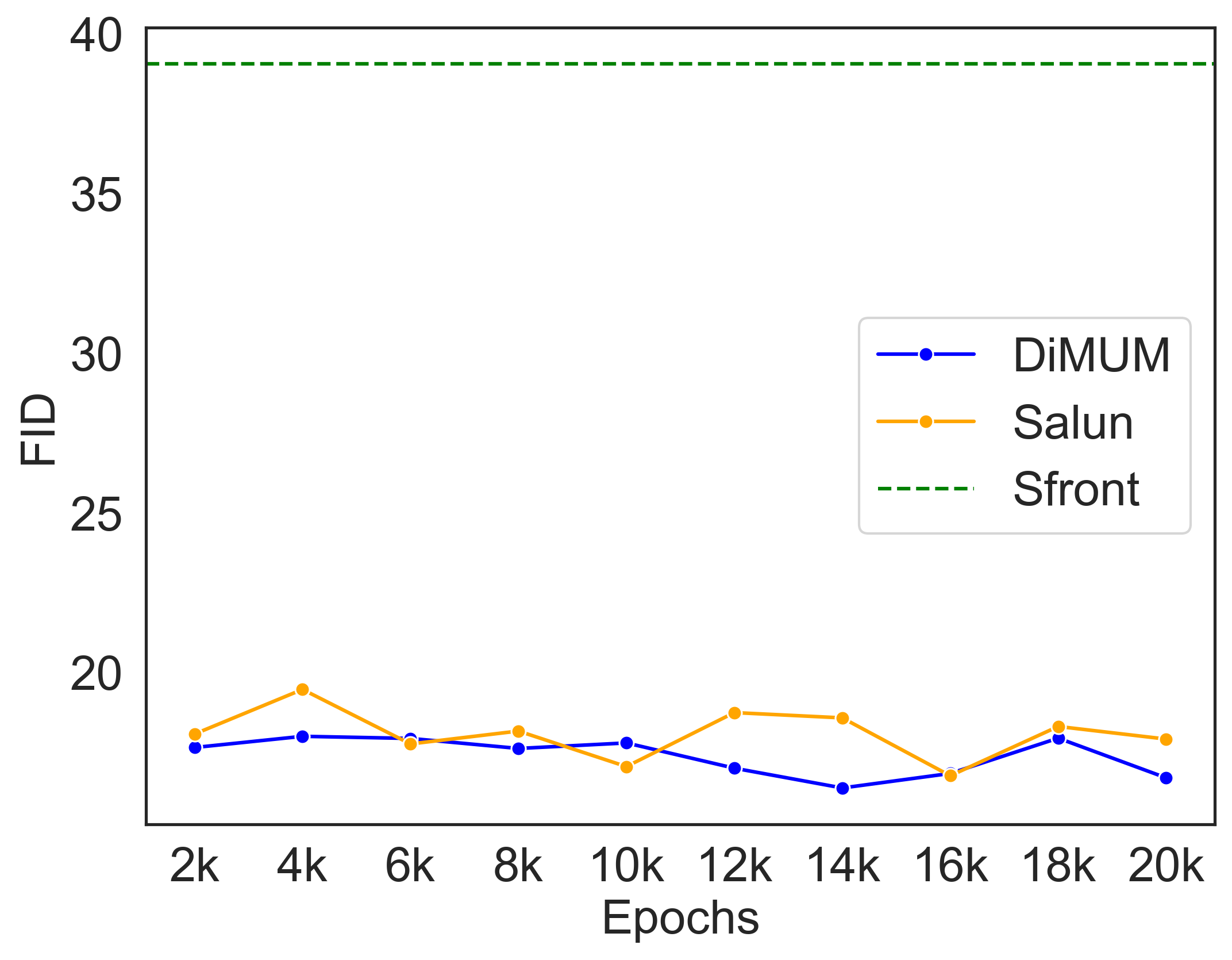}\label{fig:Unlearn_Class_1_FID}}
  \subfloat[Bird]
  {\includegraphics[width=0.2\linewidth]{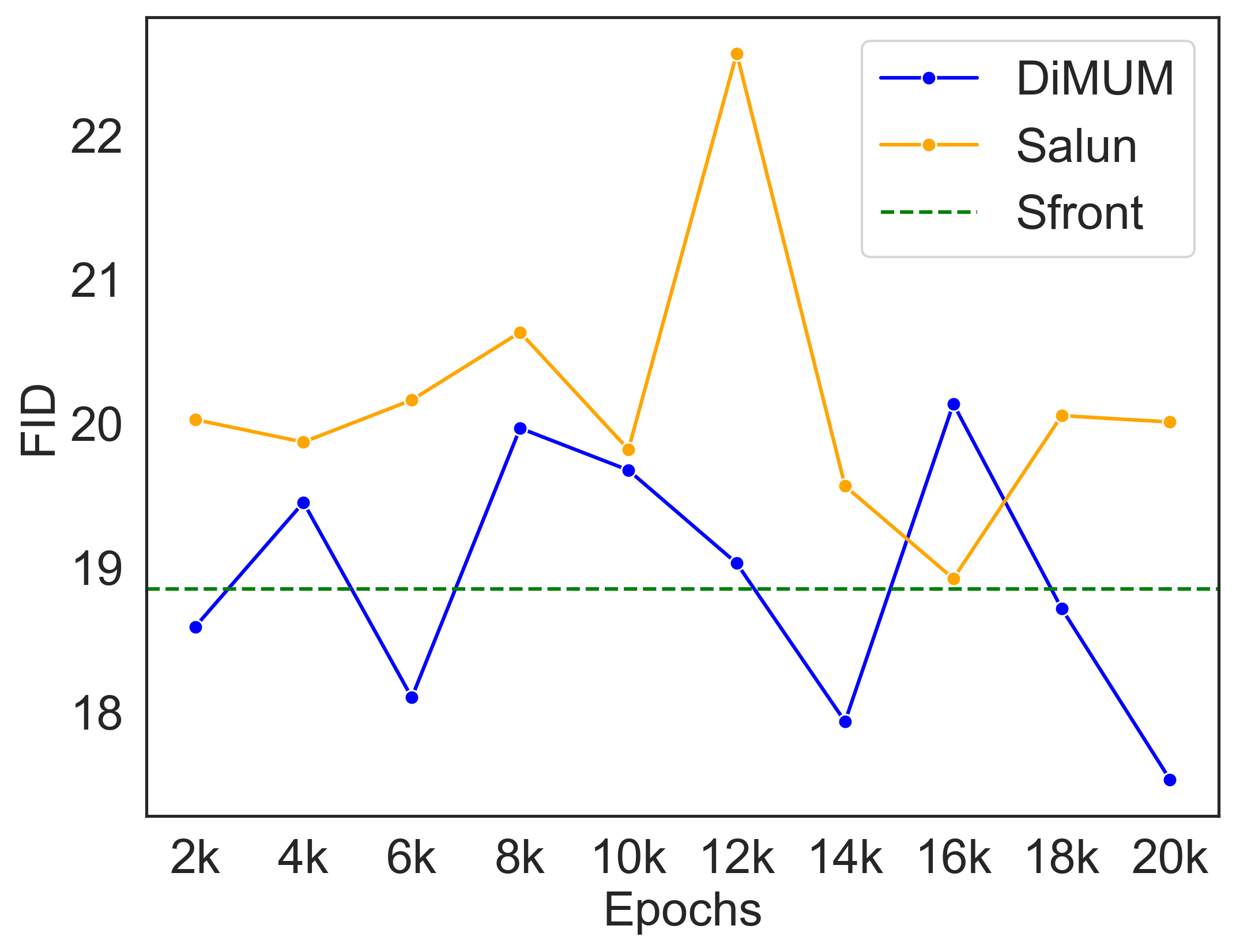}\label{fig:Unlearn_Class_2_FID}}     
  \subfloat[Cat]
  {\includegraphics[width=0.2\linewidth]{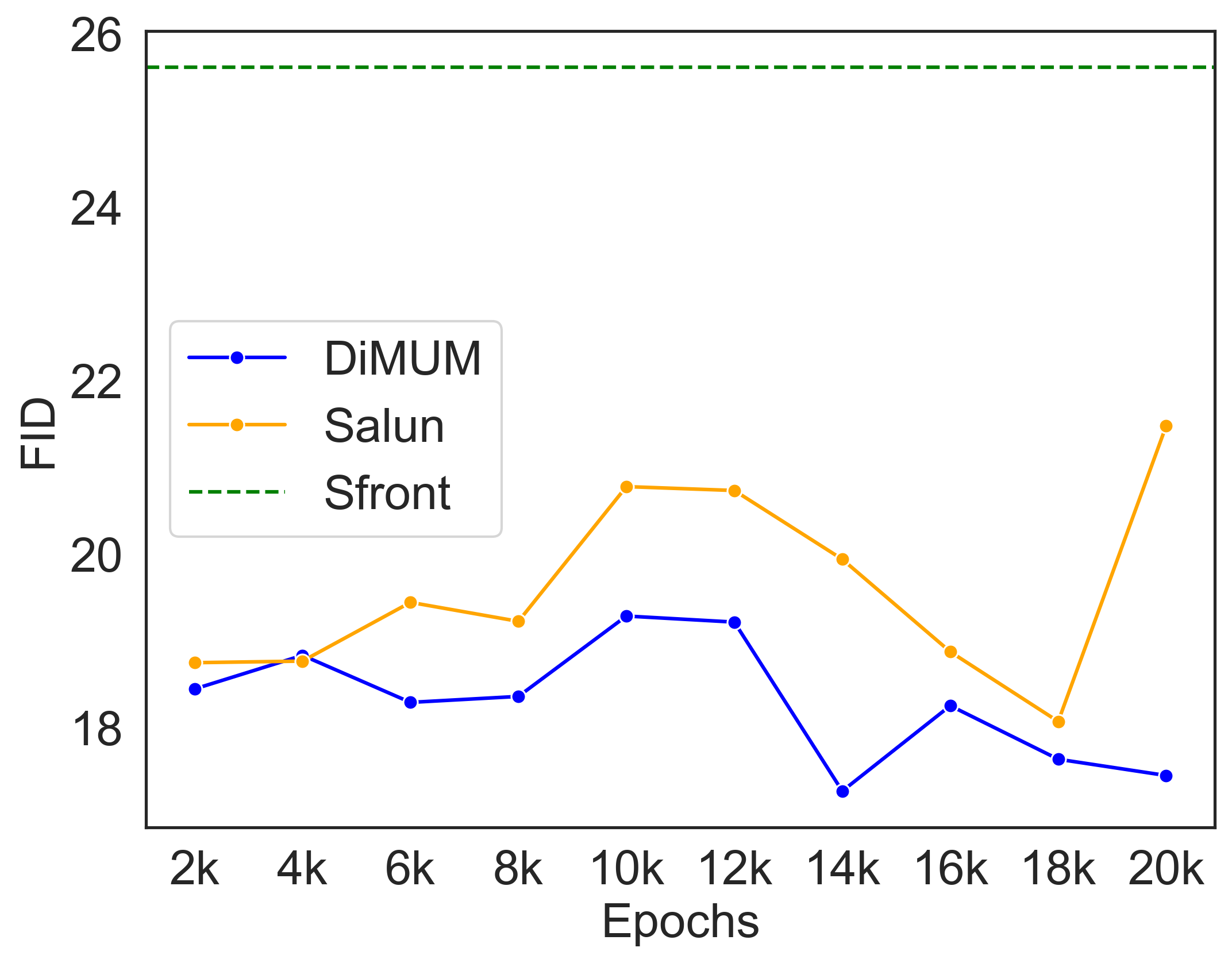}\label{fig:Unlearn_Class_3_FID}}
  \subfloat[Deer]
  {\includegraphics[width=0.2\linewidth]{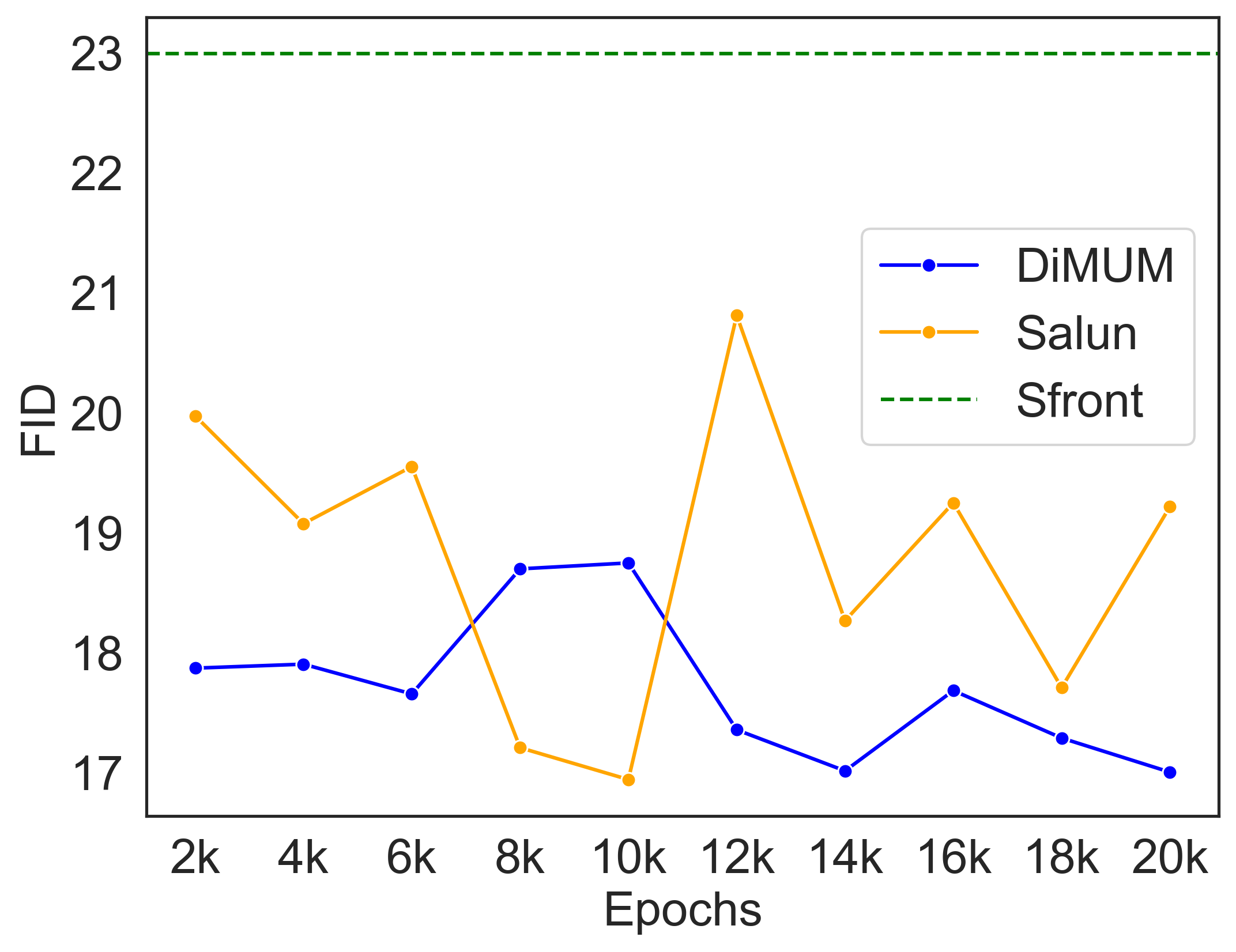}\label{fig:Unlearn_Class_4_FID}}
  \caption{This figure shows the FID curves during the unlearning process of DiMUM and Salun. We only show the FID value after 2000 unlearning steps of Sfront since it rapidly and significantly decrease the generative ability of the CDM.}
  \label{fig:fid_curves_cifar}
\end{figure*}

\subsection{Experiments on Object Unlearning} \label{sec:Unlearning a Class}

\subsubsection{Experiment Setup.}

\textbf{Base Model:}
In this section, we apply DDPM with classifier-free guidance \cite{ho2022classifier} as the base model, which is a well-known conditional DDPM whose conditioning input is the class label.
\textbf{Dataset:} The pre-trained CDM is trained on the CIFAR-10 training dataset, which consists of ten classes with 5K images per class. The image resolution is 32$\times$32. The CIFAR-10 test dataset, on the other hand, contains ten classes with 1K images per class, following a distribution similar to that of the training dataset. Thus, the test dataset can serve as the auxiliary dataset for launching DiMRA, simulating an attacker with limited background knowledge, i.e., Assumption (ii) introduced in Section \ref{sec:Attacker Model}.
\textbf{MU baselines:} 
We apply two state-of-the-art finetuning-based MU methods for CDMs, Salun \cite{fan2023salun} and Sfront \cite{huang2024unified}, as baselines. Both methods are capable of unlearning a specific class based on the class label and thus are suitable for the CIFAR-10 dataset. The loss function of Sfront, which applies the mechanism \textbf{Eq.}\eqref{eq:unlearn_loss_1}, maximizes the original CDM loss when the input is sampled from the unlearning set. On the other hand, the loss function of Salun, which applies mechanism \textbf{Eq.}\eqref{eq:unlearn_loss_2}, replaces the images of the unlearning class with images of an alternative class. For Salun and DiMUM, the alternative class is simply selected as the next class of the unlearning class in the CIFAR-10 dataset, the same as \cite{fan2023salun}.

\textbf{Evaluation process:} 
First of all, the pre-trained CDM $\boldsymbol\theta_p$ is trained on CIFAR-10 training dataset $D$ by minimizing the loss function (\ref{eq:ori_loss_c}) for 1M steps with the mini-batch size 128.
Second, we select one class of the CIFAR-10 dataset as the unlearning class (object). The unlearning dataset $D_u$ is constructed with the images of the CIFAR-10 training dataset that belong to the unlearning class. The retain dataset $D_r$ is then constructed with the remaining images of the CIFAR-10 training dataset. Next, we input $D_u$, $D_r$, and $\boldsymbol\theta_p$ into a MU method and obtain an unlearned CDM $\boldsymbol\theta_u$. To simulate an attacker with strong background knowledge, i.e., Assumption (i) in Section \ref{sec:Attacker Model}, we use the retain set as the auxiliary dataset $D_{au}$ to launch DiMRA. To simulate an attacker with limited background knowledge, i.e., Assumption (ii) in Section \ref{sec:Attacker Model}, we use the CIFAR-10 test dataset excluding the unlearning class images as the auxiliary dataset $D_{au}$ to launch DiMRA. 

\textbf{Evaluation Metrics:} Fréchet Inception Distance (FID) \cite{heusel2017gans} and Structured Fréchet Inception Distance (sFID) \cite{dhariwal2021diffusion} are commonly used to assess the distributional similarity between two image datasets. We compute FID and sFID between the synthetic dataset generated by the unlearned CDM and the retain set to assess the MU method's ability to preserve the generative capability of the pre-trained CDM, corresponding to objective (i) in Definition \ref{def:obj}. Lower FID and sFID indicate better performance with respect to objective (i). \textbf{Note} that we do not consider the unlearning class when calculating FID and sFID.
Next, we generate 1000 images with the unlearned CDM, conditioned on the label of the unlearning class. Following \cite{fan2023salun}, we use a pre-trained classification model to classify these 1000 images and compute the accuracy rate, which is the proportion of the 1000 images classified as the unlearning class. This accuracy rate is used to evaluate objective (ii) in Definition \ref{def:obj}, termed as $\text{AR}_\text{MU}$. Lower $\text{AR}_\text{MU}$ indicates better unlearning performance.
Last, we use DiMRA to attack the unlearned CDM and obtain the attacked CDM. We generate 1000 images with the attacked CDM, conditioned on the label of the unlearning class, and compute the accuracy rate with the same pre-trained classification model. This accuracy rate is used to evaluate effectiveness of DiMRA and robustness of the MU method against DiMRA, termed as $\text{AR}_\text{DiMRA}$. Lower $\text{AR}_\text{DiMRA}$ indicates more robustness against DiMRA while higher $\text{AR}_\text{DiMRA}$ indicates better attack performance.

\begin{figure} 
  \centering
  \subfloat[Pre-Trained CDM]
  {\includegraphics[width=0.48\linewidth]{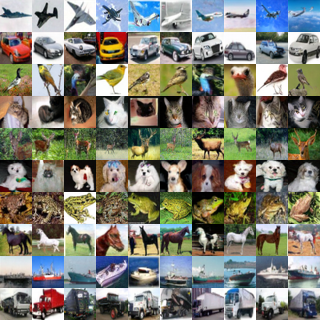}\label{fig:cifar_10_pretrain}}   \hspace{0.05cm}  
  \subfloat[Unlearned by Sfront (2050 steps)]
  {\includegraphics[width=0.48\linewidth]{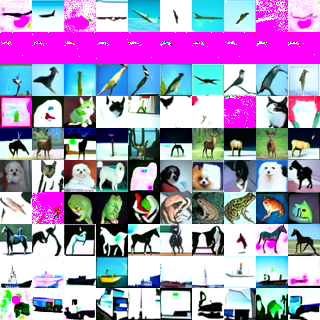}\label{fig:cifar_10_sfront}} \\
  \subfloat[Unlearned by Salun (20k steps)]
  {\includegraphics[width=0.48\linewidth]{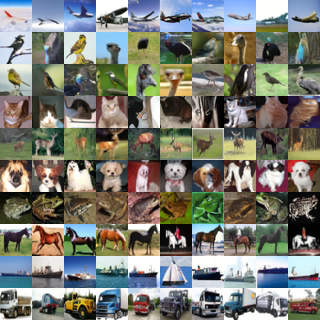}\label{fig:cifar_10_salun}}     \hspace{0.05cm}
  \subfloat[Unlearned by DiMUM (20k steps)]
  {\includegraphics[width=0.48\linewidth]{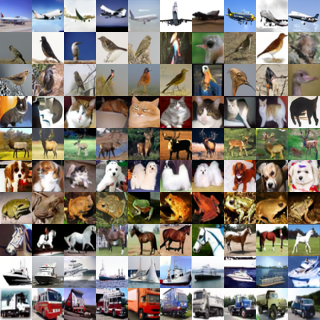}\label{fig:cifar_10_dimum}}
  \caption{This figure shows synthetic images generated by the pre-trained CDM and CDMs unlearned by Sfront, Salun, and DiMUM. The unlearning object is automobile.}
  \label{fig:synthetic_data_cifar}
\end{figure}

\subsubsection{Evaluation of Objective (i) in Definition \ref{def:obj}} \label{sec:exp_cifar_fid} 
First, we use DiMUM and Salun to unlearn the `automobile' class from the pre-trained CDM $\boldsymbol\theta_p$ for 20K optimization steps but use Sfront to unlearn the `automobile' class for only 2K optimization steps, as Sfront significantly and rapidly decreases the generative performance of the CDM during unlearning. 
Figure \ref{fig:synthetic_data_cifar} illustrates the synthetic images generated by the pre-trained CDM, along with those generated by CDMs unlearned by Sfront, Salun, and DiMUM where the unlearning class is `automobile'. As shown in Fig. \ref{fig:synthetic_data_cifar}, all three MU methods can prevent the generation of automobile images. However, the images generated by the CDM unlearned with Sfront become unrecognizable at 2050 unlearning steps due to the divergence of its unlearning loss and thus fails to achieve objective (i). On the other hand, the CDMs unlearned by Salun and DiMUM generate bird images given the conditioning input `automobile' and present similar generative performance to the pre-trained CDM when generating other classes.
Then, we repeat the above experiments to unlearn another four classes, i.e., airplane, bird, cat, and deer of the CIFAR-10 dataset. Figure. \ref{fig:fid_curves_cifar} shows the FID curves during the unlearning process and the results are consistent among the five unlearning classes. 
As illustrated, DiMUM achieves the highest generative performance after unlearning. Sfront exhibits the worst generative performance, despite being conducted for only 2K steps. DiMUM's superior generative performance after unlearning can be attributed to the convergence of both its unlearning and retain losses. This joint optimization enables the CDM's parameters to reach a new local optimum with respect to the retain loss and thus preserves the generative performance given the conditioning input of the retain set.
Table \ref{tab:fid-cifar} summarizes the generative performance of the CDM after the last unlearning step, i.e., 2K for Sfront and 20K for DiMUM and Salun.

\begin{table}
    \centering
    \caption{FID, sFID, and AR$_\text{MU}$ after Unlearning}
    \begin{tabular}{ccccc}
        \toprule
        \centering
        \textbf{Unlearn class} & \textbf{Method} & \textbf{FID} $\downarrow$ & \textbf{sFID} $\downarrow$ & \textbf{AR$_\text{MU}$} $\downarrow$ \\
        \midrule
        \centering
        \multirow{3}*{\textbf{Airplane}} & Salun\cite{fan2023salun} & 16.0395 & 30.2921 & 0 \\
        \centering {} & Sfront\cite{huang2024unified} & 17.5918 & 32.0599 & 0  \\
        \centering {} & \textbf{\textcolor{blue}{DiMUM (Ours)}} & \textbf{14.6809} & \textbf{30.0606} & 0  \\
        \midrule
        \centering \multirow{3}*{\textbf{Automobile}} & Salun\cite{fan2023salun} & 17.9248 & 32.8567 & 0 \\
        \centering {} & Sfront\cite{huang2024unified} & 39.0697 & 44.3232 & 0   \\
        \centering {} & \textbf{\textcolor{blue}{DiMUM (Ours)}} & \textbf{16.7114} & \textbf{32.1452} & 0  \\
        \midrule
        \centering \multirow{3}*{\textbf{Bird}} & Salun\cite{fan2023salun} & 20.0131 & 33.5968 & 0\\
        \centering {} & Sfront\cite{huang2024unified} & 18.8547 & 33.5163 & 0 \\
        \centering {} & \textbf{\textcolor{blue}{DiMUM (Ours)}} & \textbf{17.5327} & \textbf{32.6799} & 0  \\
        \midrule
        \centering \multirow{3}*{\textbf{Cat}} & Salun\cite{fan2023salun} & 21.4828 & 34.4921 & 0  \\
        \centering {} & Sfront\cite{huang2024unified} & 25.6169 & 36.4590 & 0 \\
        \centering {} & \textbf{\textcolor{blue}{DiMUM (Ours)}} & \textbf{17.4537} & \textbf{32.6307} & 0 \\
        \midrule
        \centering \multirow{3}*{\textbf{Deer}} & Salun\cite{fan2023salun} & 19.2202 & 32.9396 & 0\\
        \centering {} & Sfront\cite{huang2024unified} & 22.9986 & 35.6444 & 0 \\
        \centering {} & \textbf{\textcolor{blue}{DiMUM (Ours)}} & \textbf{17.0063} & \textbf{32.2369} & 0  \\
        \bottomrule
    \end{tabular}
    \label{tab:fid-cifar}
\end{table}

\begin{table}
\centering
\caption{Largest AR$_{\text{DiMRA}}$ during Attack for Different Methods, Unlearning Steps and Labels with Assumption (i)}
\begin{tabular}{p{1.3cm}p{0.7cm}p{0.8cm}p{0.8cm}p{0.8cm}p{0.8cm}p{0.8cm}}
\hline
\textbf{Method} & \textbf{Steps} & \textbf{Plane} & \textbf{Auto} & \textbf{Bird} & \textbf{Cat} & \textbf{Deer} \\
\hline
\multirow{2}*{{Sfront}\cite{huang2024unified}} & 1K  & 0.996 & 1 & 1 & 1 & 1 \\
{} & 2K  & 0.974 & 0.960 & 0.999 & 0.988 & 1 \\
\midrule
\multirow{2}*{{Salun}\cite{fan2023salun}}  & 10K & 0.647 & 0.747 & 0.807 & 0.916 & 0.866 \\
{}  & 20K & 0.187 & 0.373 & 0.537 & 0.813 & 0.657 \\
\midrule
\multirow{2}*{{\textbf{\textcolor{blue}{DiMUM }}}}  & 10K & 0.260 & 0.414 & 0.566 & 0.719 & 0.627 \\
{}  & 20K & \textbf{0.028} & \textbf{0.032} & \textbf{0.166} & \textbf{0.252} & \textbf{0.217} \\
\hline
\end{tabular}
\label{tab:ar_cifar_strong_assumption}
\end{table}

\begin{table}
\centering
\caption{Largest AR$_{\text{DiMRA}}$ during Attack for Different Methods, Unlearning Steps and Labels with Assumption (ii)}
\begin{tabular}{p{1.3cm}p{0.7cm}p{0.8cm}p{0.8cm}p{0.8cm}p{0.8cm}p{0.8cm}}
\hline
\textbf{Method} & \textbf{Steps} & \textbf{Plane} & \textbf{Auto} & \textbf{Bird} & \textbf{Cat} & \textbf{Deer} \\
\hline
\multirow{2}*{{Sfront}\cite{huang2024unified}} & 1K  & 0.908 & 0.918 & 0.980 & 0.959 & 0.997 \\
{} & 2K  & 0.366 & 0.642 & 0.970 & 0.920 & 0.997 \\
\midrule
\multirow{2}*{{Salun}\cite{fan2023salun}}   & 10K & 0.277 & 0.452 & 0.501 & 0.670 & 0.718 \\
{}  & 20K & 0.042 & 0.135 & 0.313 & 0.691 & 0.456 \\
\midrule
\multirow{2}*{{\textbf{\textcolor{blue}{DiMUM}}}} & 10K & 0.060 & 0.259 & 0.278 & 0.489 & 0.429 \\
{}  & 20K & \textbf{0.010} & \textbf{0.007} & \textbf{0.055} & \textbf{0.204} & \textbf{0.117} \\
\hline
\end{tabular}
\label{tab:ar_cifar_weak_assumption}
\end{table}

\begin{figure} 
  \centering
  \subfloat[CDM unlearned by Sfront]
  {\includegraphics[width=0.48\linewidth]{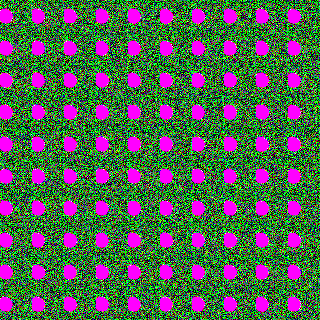}}   \hspace{0.05cm}  
  \subfloat[Attacked CDM (Sfront)]
  {\includegraphics[width=0.48\linewidth]{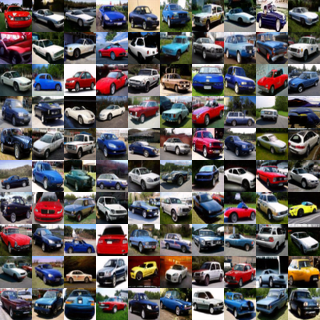}} \\
  \subfloat[CDM unlearned by Salun]
  {\includegraphics[width=0.48\linewidth]{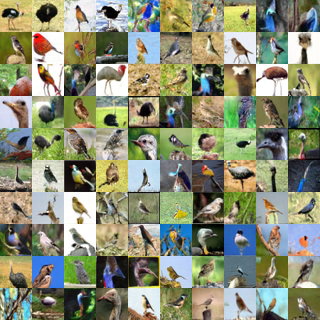}}     \hspace{0.05cm}
  \subfloat[Attacked CDM (Salun)]
  {\includegraphics[width=0.48\linewidth]{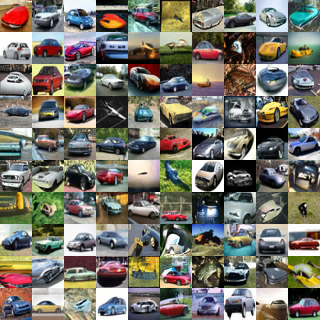}} \\
  \subfloat[CDM unlearned by DiMUM]
  {\includegraphics[width=0.48\linewidth]{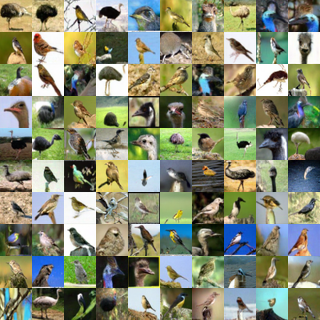}}     \hspace{0.05cm}
  \subfloat[Attacked CDM (DiMUM)]
  {\includegraphics[width=0.48\linewidth]{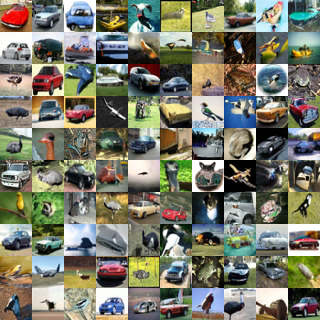}}
  \caption{The left side and right side of this figure show synthetic images generated by unlearned CDMs and corresponding attacked CDMs with the conditioning input `automobile'.}
  \label{fig:synthetic_data_cifar_attack}
  \vspace{-3mm}
\end{figure}

\begin{figure*} 
  \centering
  \subfloat[Airplane]
  {\includegraphics[width=0.2\linewidth]{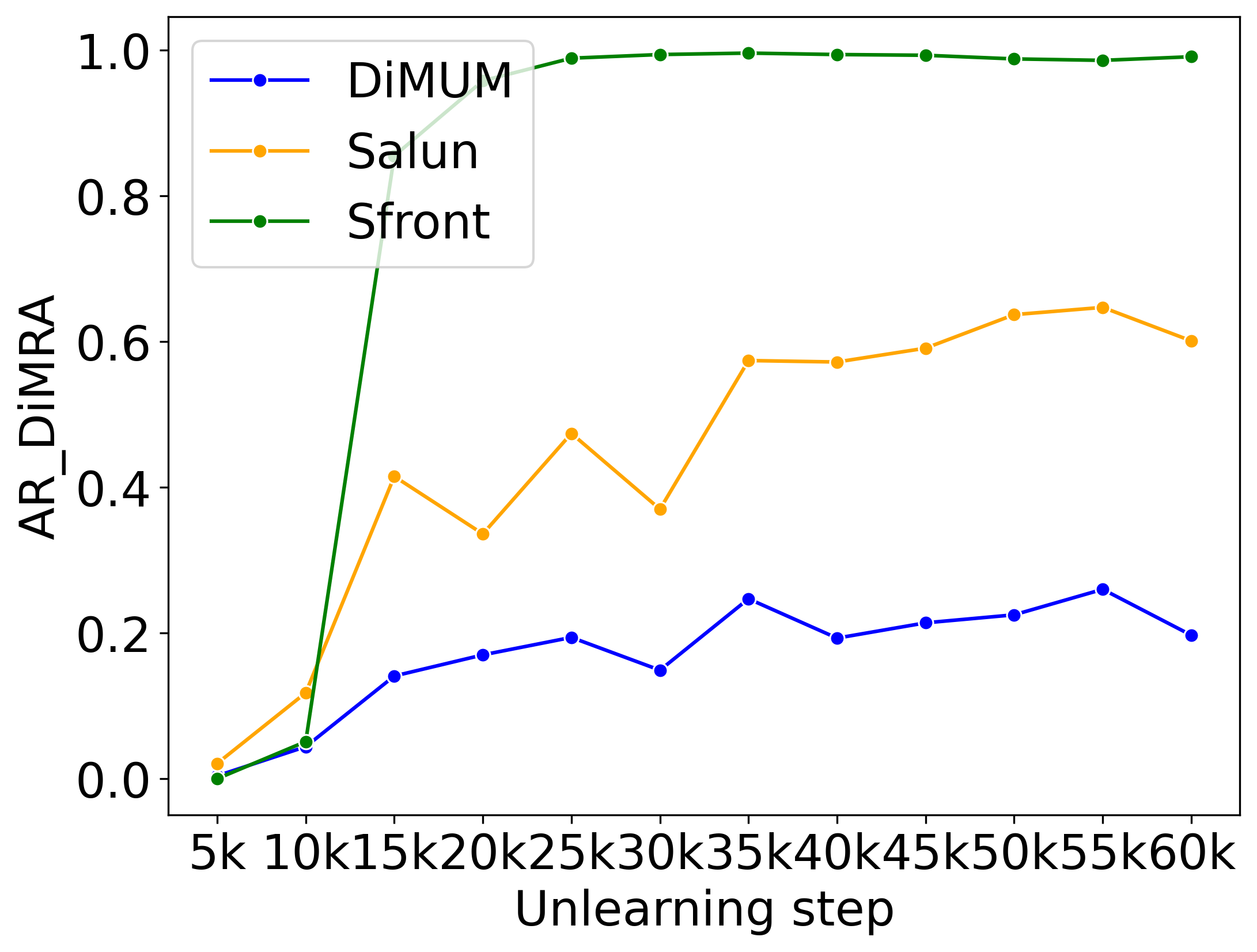}\label{fig:attack_Class_0_ar_10k}}     
  \subfloat[Automobile]
  {\includegraphics[width=0.2\linewidth]{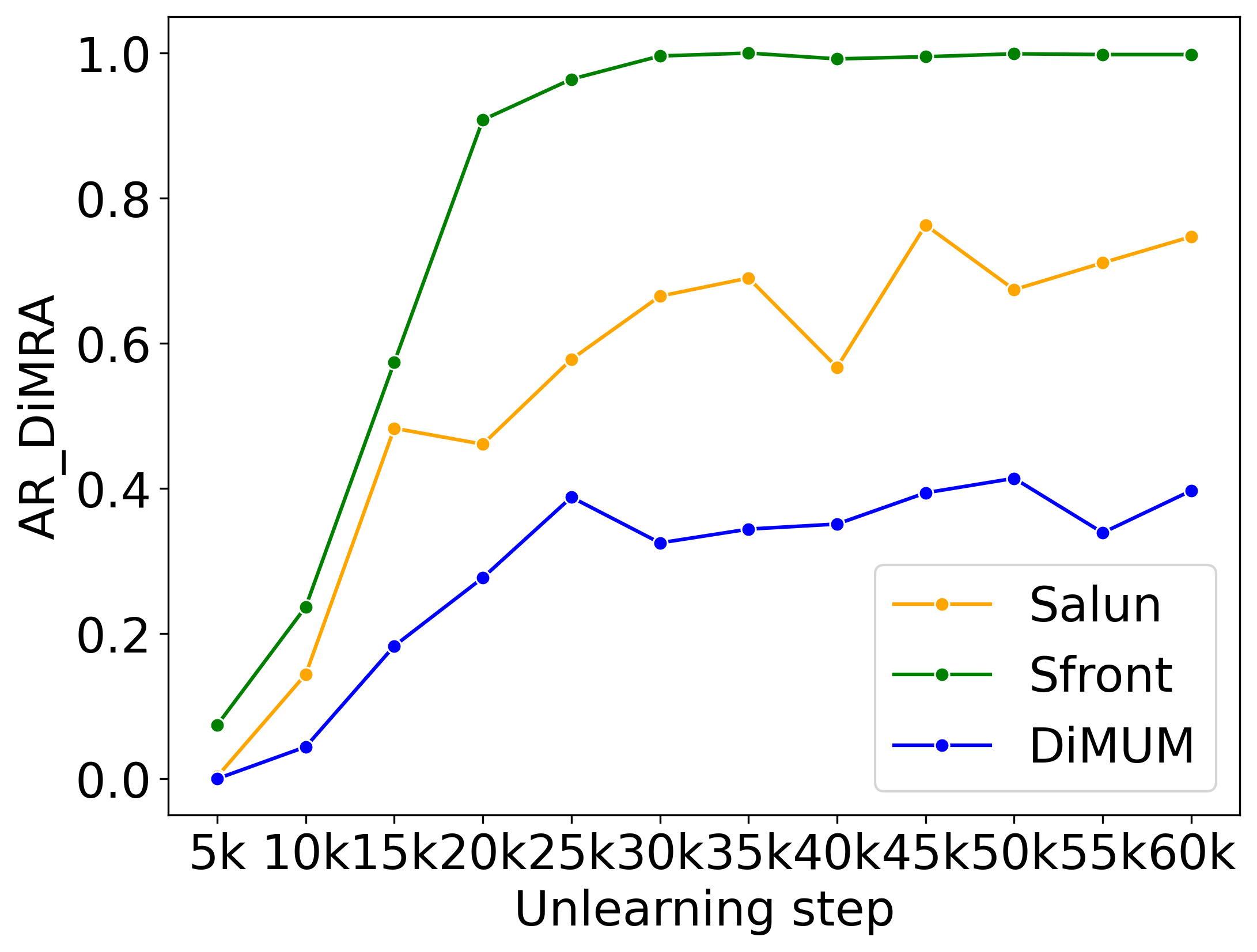}\label{fig:attack_Class_1_ar_10k}}
  \subfloat[Bird]
  {\includegraphics[width=0.2\linewidth]{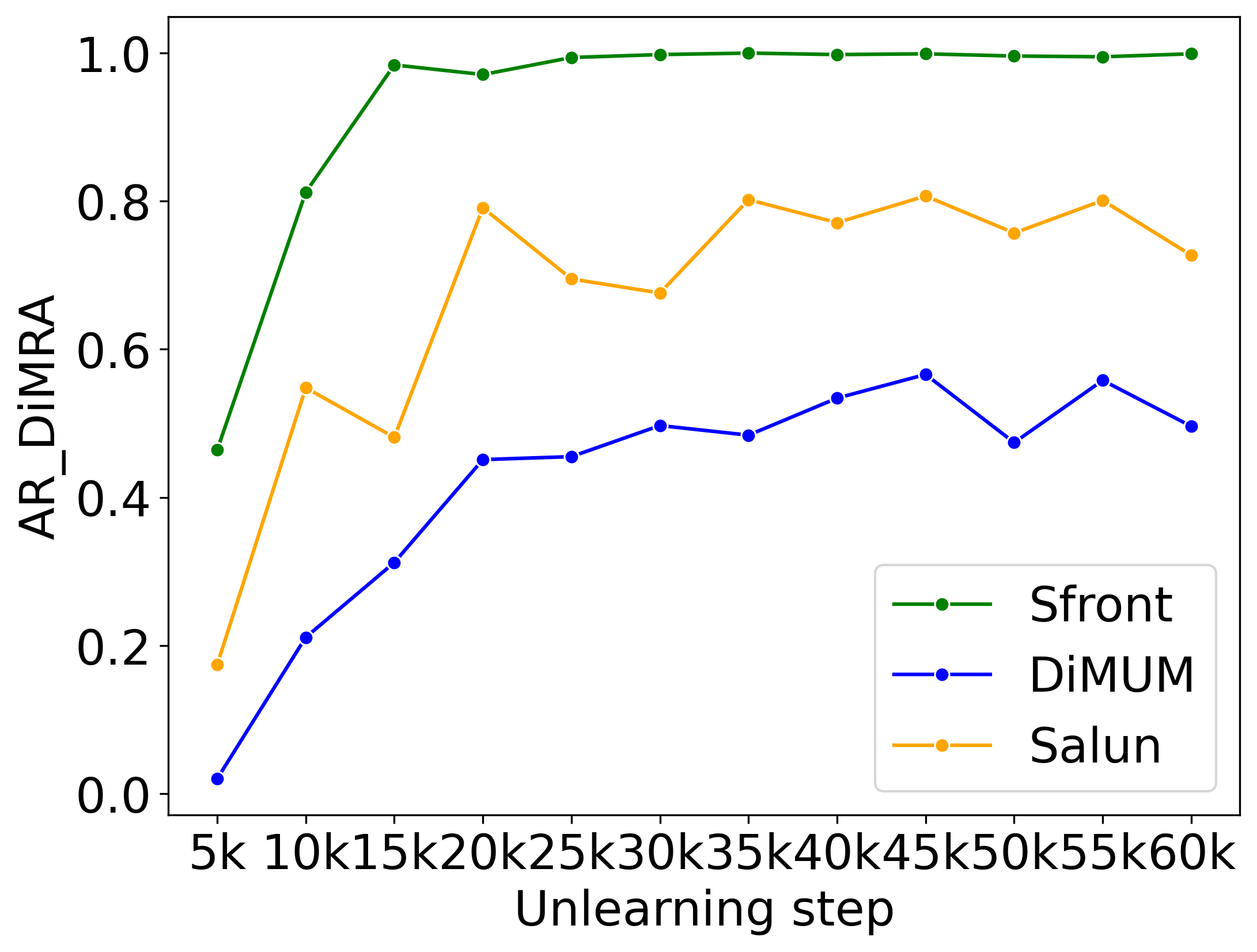}\label{fig:attack_Class_2_ar_10k}}     
  \subfloat[Cat]
  {\includegraphics[width=0.2\linewidth]{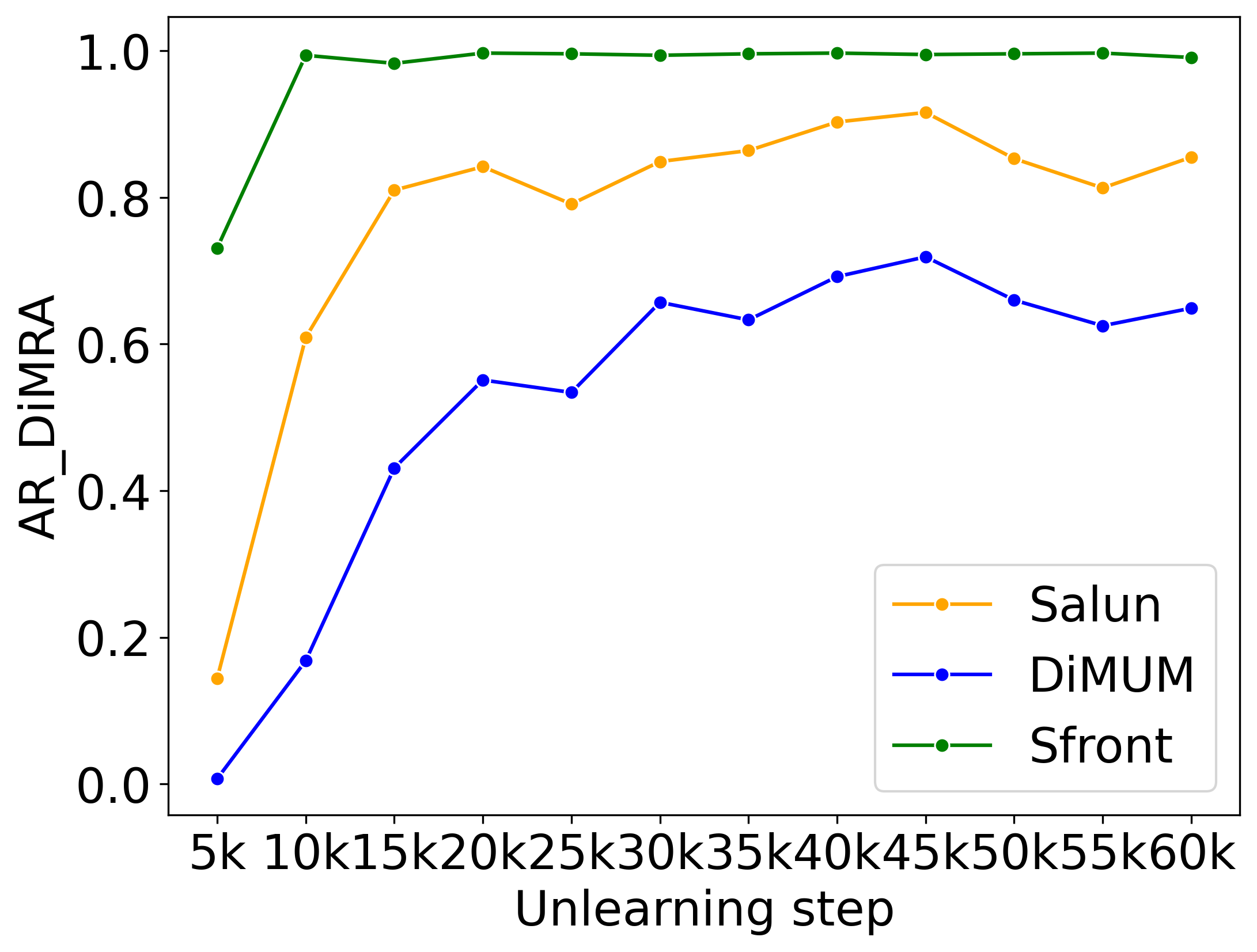}\label{fig:attack_Class_3_ar_10k}}
  \subfloat[Deer]
  {\includegraphics[width=0.2\linewidth]{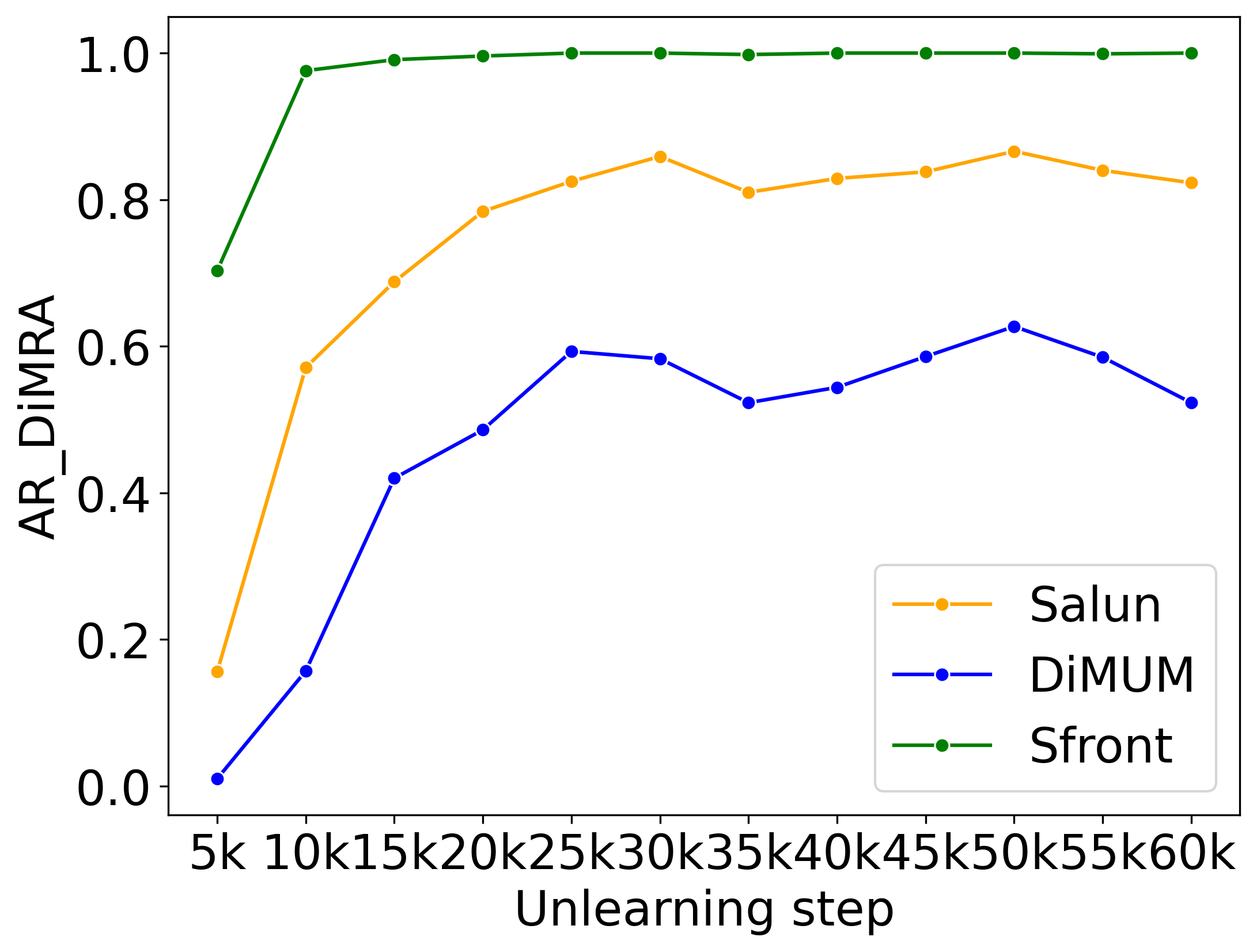}\label{fig:attack_Class_4_ar_10k}}
  \caption{This figure shows the AR$_{\text{DiMRA}}$ curves during the attack process of DiMRA where the CDMs are unlearned for 10K steps by Salun and DiMUM and 1K steps by Sfront for five unlearning classes.}
  \label{fig:ar_curves_cifar_10k}
  \vspace{-3mm}
\end{figure*}

\begin{figure*} 
  \centering
  \subfloat[Airplane]
  {\includegraphics[width=0.2\linewidth]{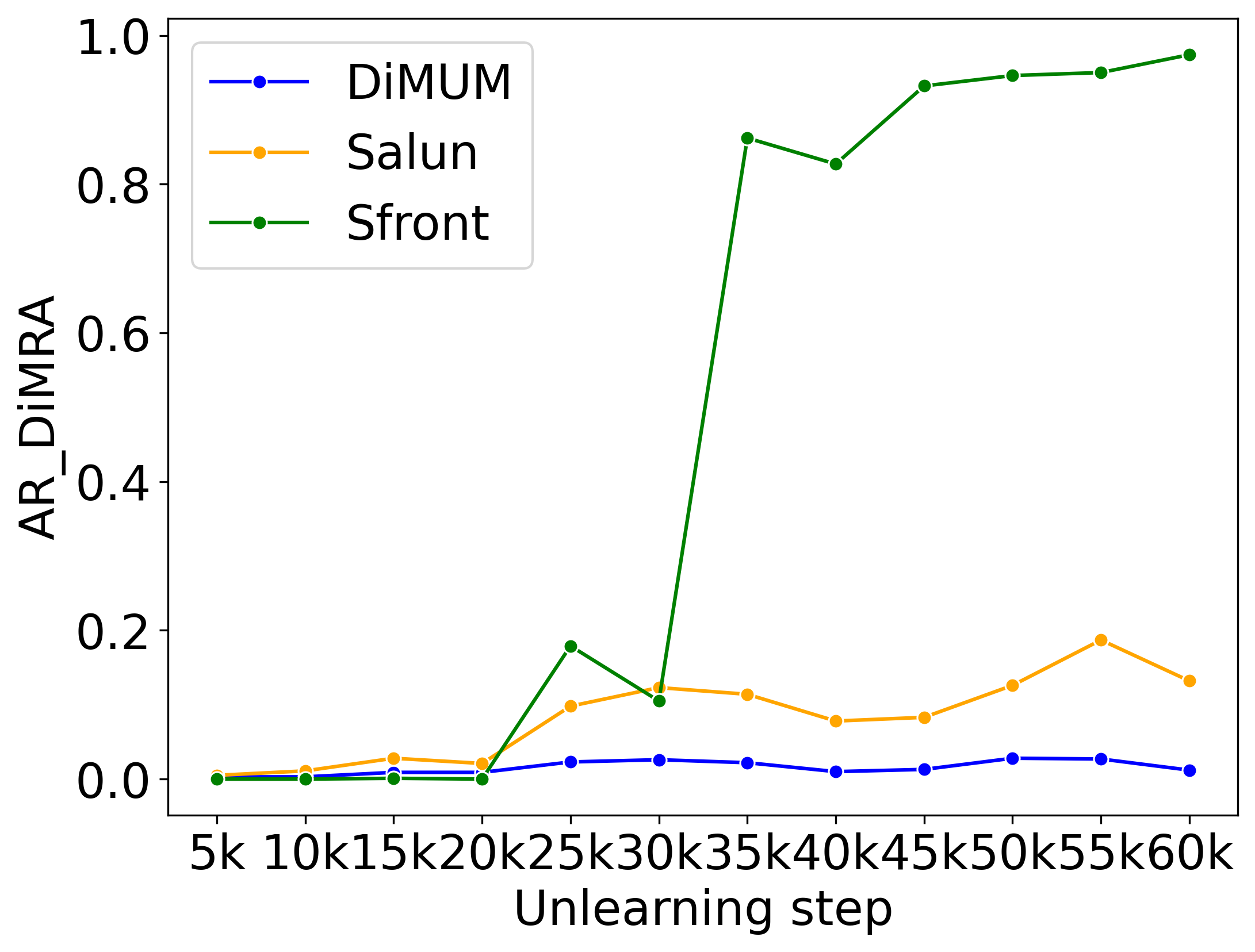}\label{fig:attack_Class_0_ar_20k}}     
  \subfloat[Automobile]
  {\includegraphics[width=0.2\linewidth]{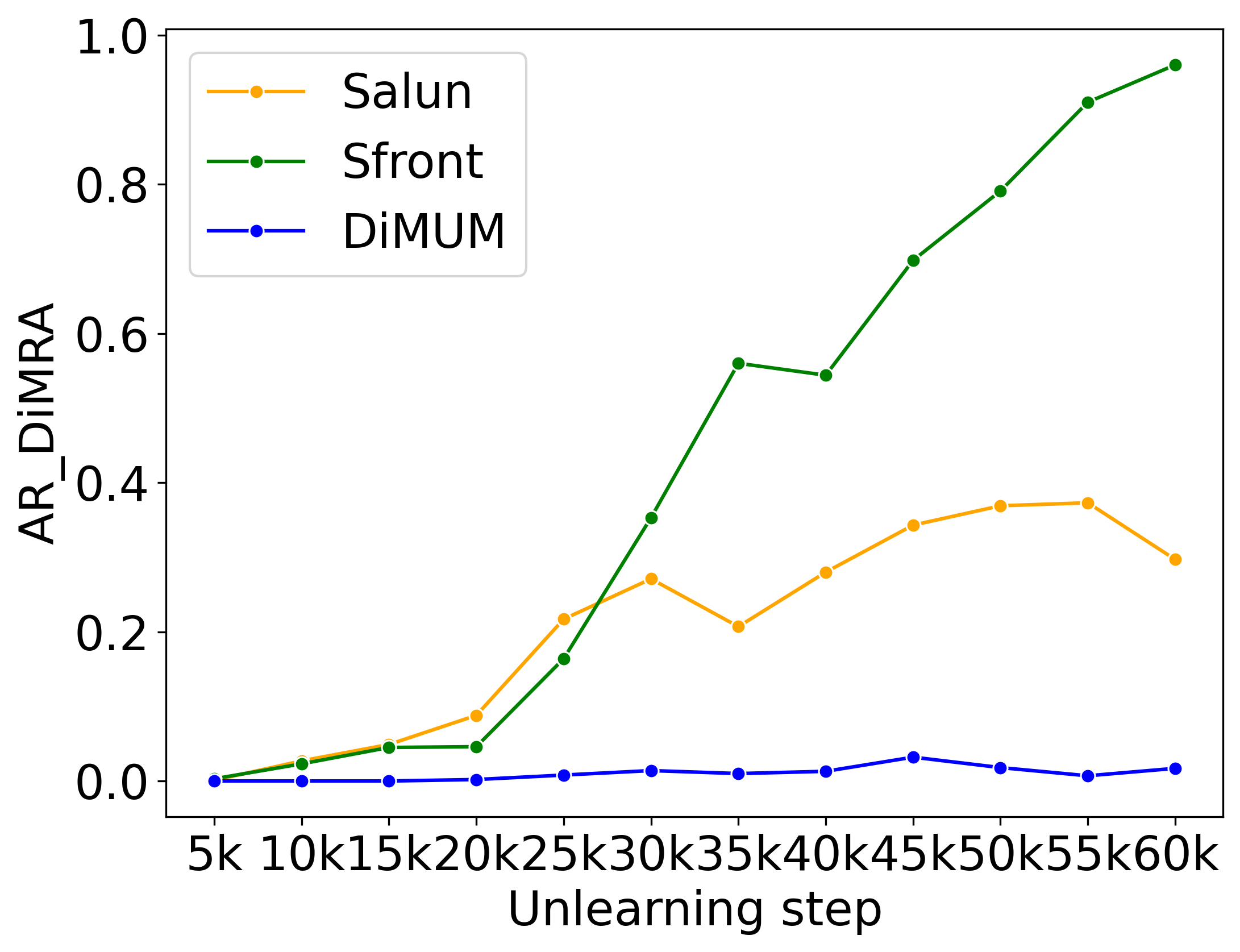}\label{fig:attack_Class_1_ar_20k}}
  \subfloat[Bird]
  {\includegraphics[width=0.2\linewidth]{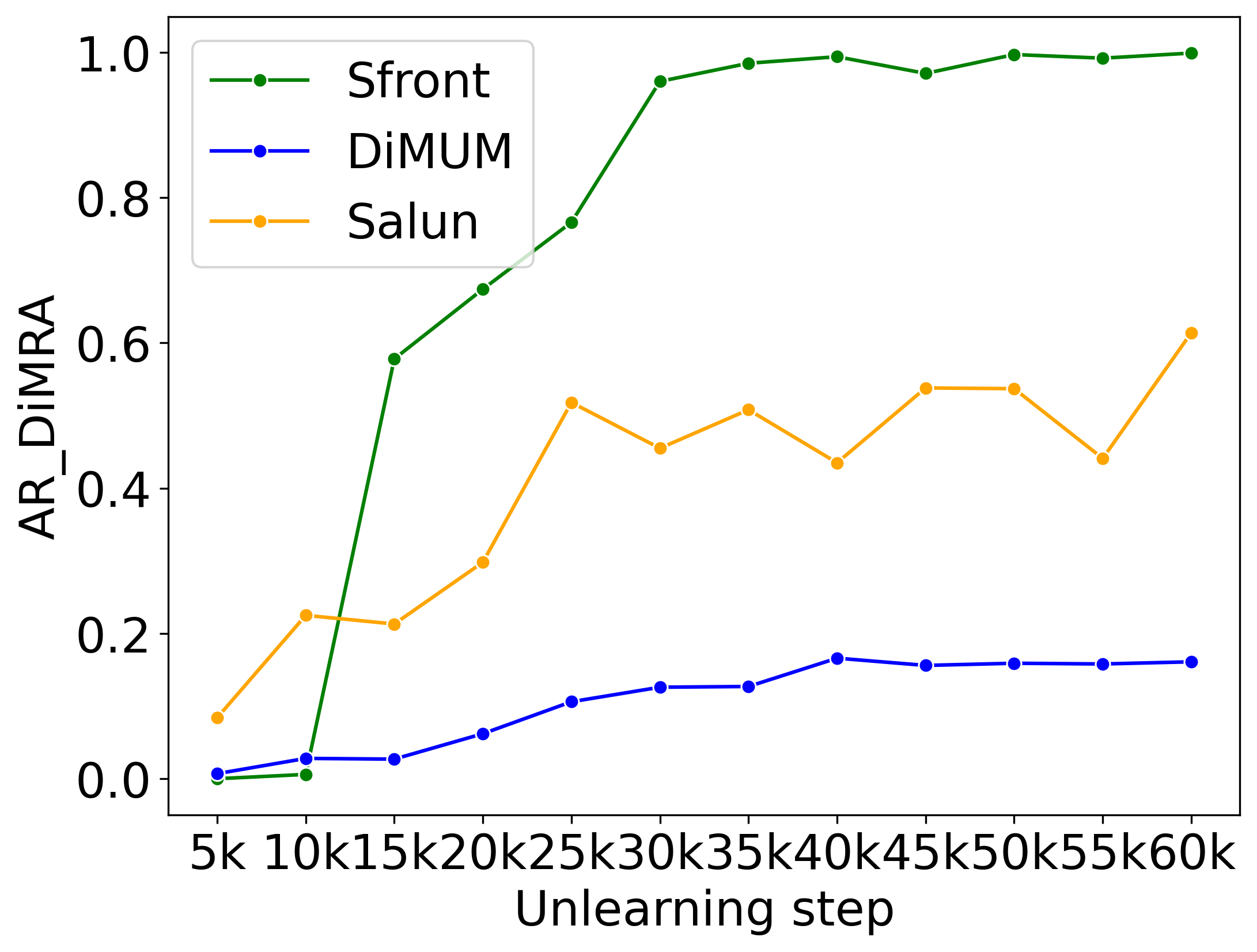}\label{fig:attack_Class_2_ar_20k}}     
  \subfloat[Cat]
  {\includegraphics[width=0.2\linewidth]{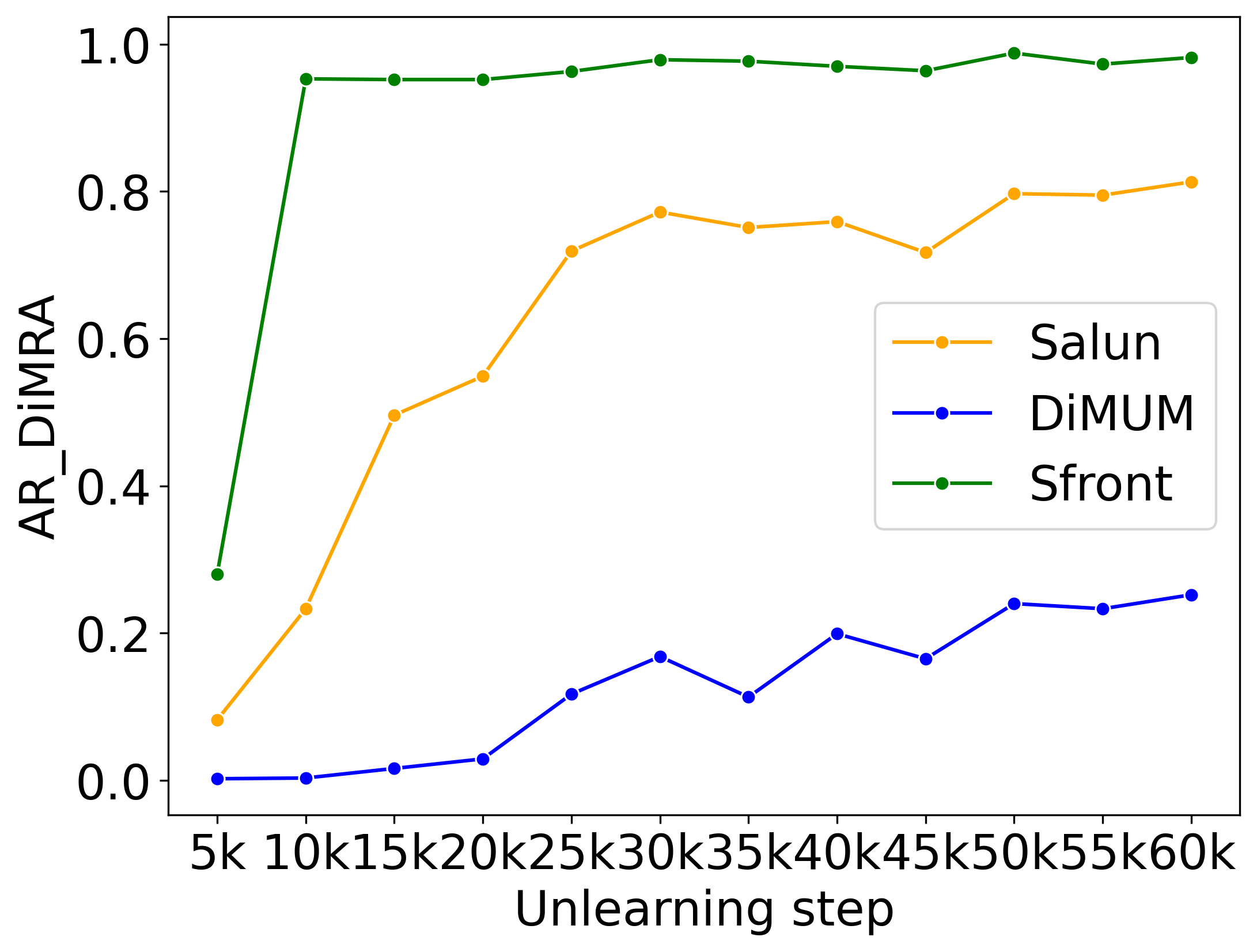}\label{fig:attack_Class_3_ar_20k}}
  \subfloat[Deer]
  {\includegraphics[width=0.2\linewidth]{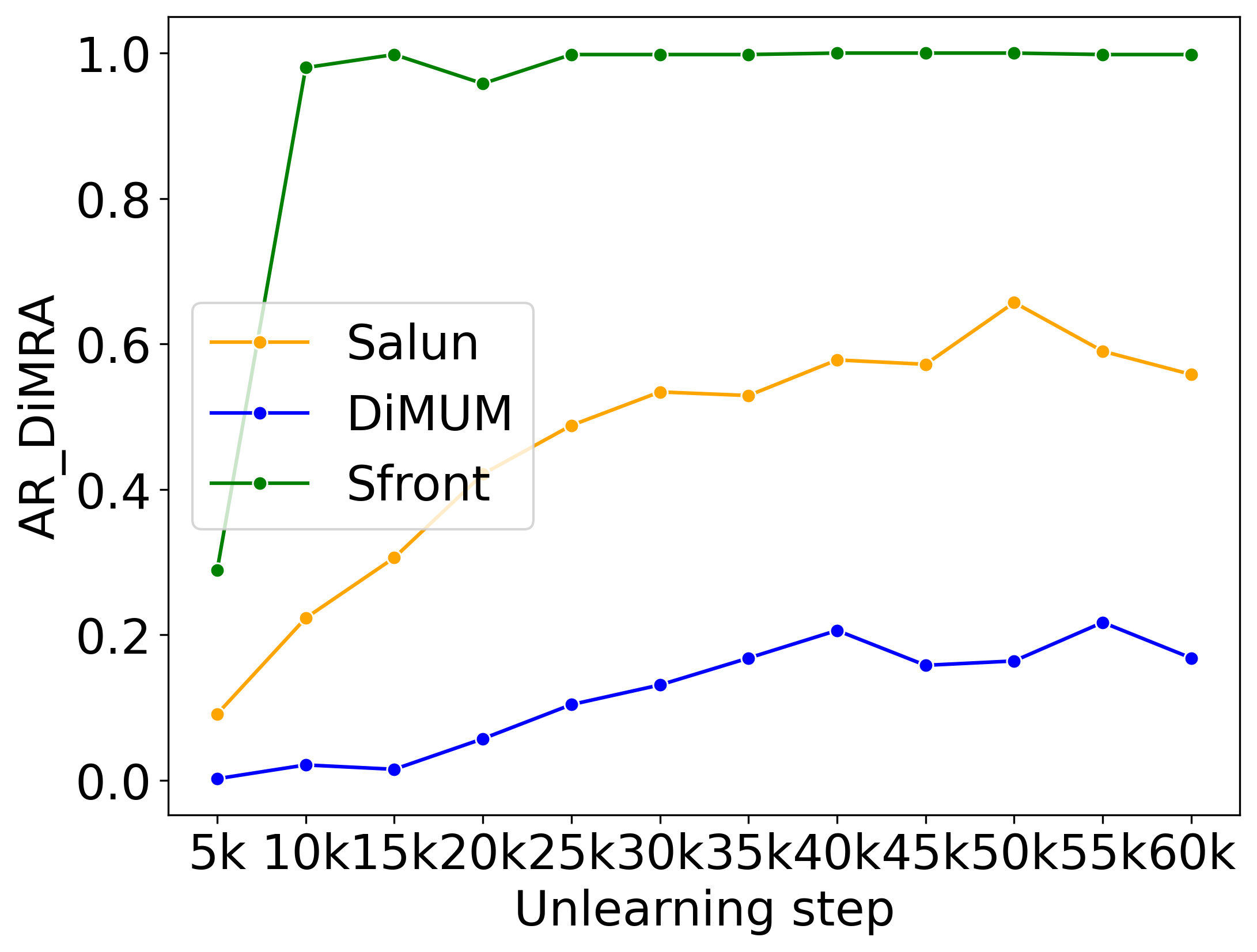}\label{fig:attack_Class_4_ar_20k}}
  \caption{This figure shows the AR$_{\text{DiMRA}}$ curves during the attack process of DiMRA where the CDMs are unlearned for 20K steps by Salun and DiMUM and 2K steps by Sfront for five unlearning classes.}
  \label{fig:ar_curves_cifar_20k}
  \vspace{-3mm}
\end{figure*}

\subsubsection{Evaluation of Objective (ii) in Definition \ref{def:obj}} 

\begin{figure} 
  \centering
  \includegraphics[width=0.7\linewidth]{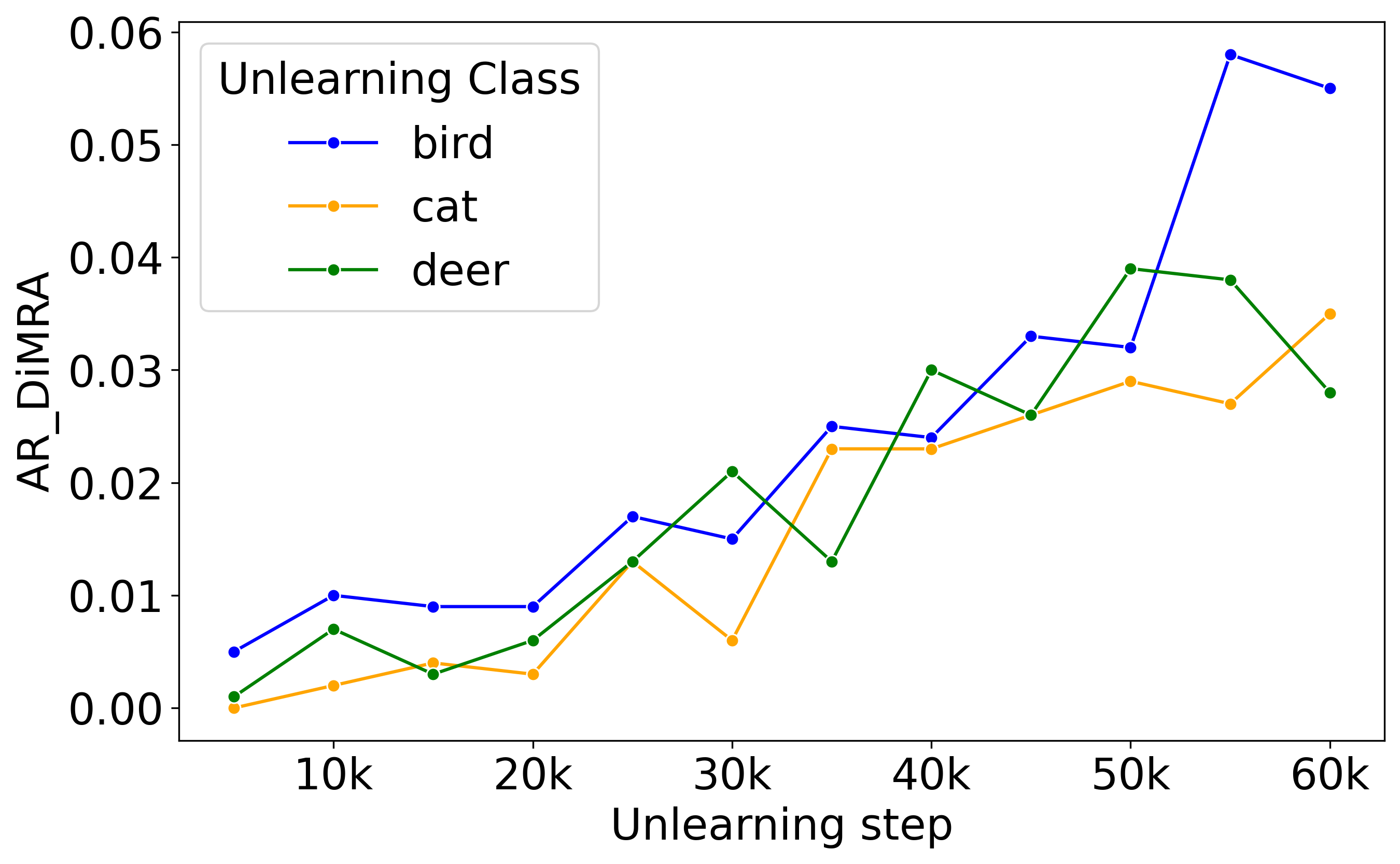}
  \caption{This figure shows the AR$_{\text{DiMRA}}$ curves during the attack process of DiMRA where the CDMs are unlearned for 40K optimization steps DiMUM.}
  \label{fig:ar_curves_cifar_40k}
  \vspace{-3mm}
\end{figure}

First, we launch DiMRA under Assumption (i) introduced in Section \ref{sec:Attacker Model}, where the auxiliary dataset is the retained set. We select CDMs that have been unlearned by Salun and DiMUM for 10K optimization steps, and by Sfront for 1K optimization steps. DiMRA is then applied to attack these unlearned CDMs for 60K optimization steps. 
Figure \ref{fig:synthetic_data_cifar_attack} presents synthetic images generated by both unlearned and attacked CDMs, with the conditioning input `automobile', where the unlearning class is `automobile'. As shown in the left side of Fig. \ref{fig:synthetic_data_cifar_attack}, after being unlearned by Sfront, Salun, and DiMUM, the CDMs do not generate automobile images, i.e., AR$_{\text{MU}}$=0 as shown in Table \ref{tab:fid-cifar}, indicating all three methods achieve objective (ii). However, as shown in the right side of Fig. \ref{fig:synthetic_data_cifar_attack}, DiMRA can effectively reverse these three MU methods. After being attacked for 60K steps, the CDM unlearned by Sfront generates high-quality `automobile' images given the conditioning input `automobile'. In contrast, DiMUM is the most robust against DiMRA since the corresponding attacked CDM generates the least `automobile' images.
Figure \ref{fig:ar_curves_cifar_10k} presents the AR$_{\text{DiMRA}}$ curves during the attack, and the results are consistent among the five unlearning classes. As illustrated, Sfront is the most vulnerable to DiMRA despite having the most significant decrease in generative performance (Fig. \ref{fig:fid_curves_cifar}). Additionally, DiMUM demonstrates greater robustness against DiMRA compared to Salun.
Then, we select CDMs unlearned by Salun and DiMUM for 20K steps and by Sfront for 2K steps and use DiMRA to attack these CDMs. Comparing Fig. \ref{fig:ar_curves_cifar_20k} and Fig. \ref{fig:ar_curves_cifar_10k}, we can conclude that longer unlearning steps improve the robustness of the unlearned CDM against DiMRA. Comparing AR$_{\text{DiMRA}}$ curves across different unlearning classes, we observe that the difficulty of unlearning varies by the unlearning class. 
For instance, after 20K unlearning steps of DiMUM, AR$_{\text{DiMRA}}$ is below 5\% when the unlearning class is `airplane' and `automobile', but exceeds 10\% when the unlearning class is `bird', `cat', and `deer'. 
Table \ref{tab:ar_cifar_strong_assumption} summarizes the highest AR$_{\text{DiMRA}}$ during the attack period for each MU method and unlearning class. 
Moreover, as shown in Fig. \ref{fig:ar_curves_cifar_40k}, after 40K unlearning steps of DiMUM, AR$_{\text{DiMRA}}$ falls below 6\% for the unlearning classes `bird', `cat', and `deer', highlighting the potential of DiMUM for irreversible MU. 

Now, we apply DiMRA to attack the unlearned CDMs under Assumption (ii) described in Section \ref{sec:Attacker Model}, where the auxiliary dataset is the CIFAR-10 test set excluding images belonging to the unlearning class. Table \ref{tab:ar_cifar_weak_assumption} summarizes the highest AR$_{\text{DiMRA}}$ during the 60K attack steps for each MU method and unlearning class. Comparing Table \ref{tab:ar_cifar_strong_assumption} and Table \ref{tab:ar_cifar_weak_assumption}, we can find that the AR$_{\text{DiMRA}}$ becomes smaller under Assumption (ii) but still remains at a high level. This indicates that DiMRA is effective under a realistic assumption. Besides, the results in Table \ref{tab:ar_cifar_weak_assumption} are consistent with those in Table \ref{tab:ar_cifar_strong_assumption}, i.e., more unlearning steps can weaken DiMRA and the proposed DiMUM is the most robust against DiMRA.


\subsection{Experiments on Feature Unlearning}

\subsubsection{Experiment Setup.}

\textbf{Base Model:}
A recent benchmark paper \cite{zhang2024unlearncanvas} about MU in CDMs has released a large-scale latent diffusion model \cite{rombach2022high}, which is obtained by finetuning the open-source Stable Diffusion (SD) v1.5 on the UnlearnCanvas dataset. In this section, we use the finetuned CDM\footnote{\url{https://github.com/OPTML-Group/UnlearnCanvas}} released in \cite{zhang2024unlearncanvas} as the pre-trained CDM.

\textbf{Dataset:} We apply UnlearnCanvas dataset introduced in \cite{zhang2024unlearncanvas}. The UnlearnCanvas contains 20 classes of objects with 50 kinds of art styles. There are 20 images with resolution 512$\times$512 for each specific combination of class and art style. The text prompts in UnlearnCanvas follow the format of `\textit{An image of \{CLASS\} in \{ART STYLE\} style.}'. The conditioning features are derived by encoding text prompts using a pre-trained text encoder, whose parameters are frozen during the unlearning and attack processes.

\begin{figure} [t]
  \vspace{-2mm}
  \centering
  \subfloat[Van Gogh style]
  {\includegraphics[width=\linewidth]{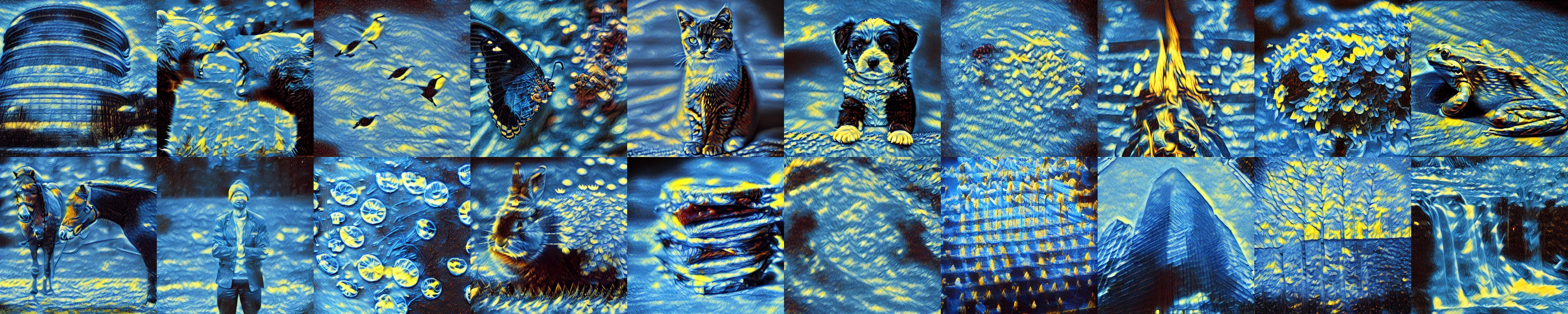}}   \\
  \subfloat[Abstractionism style]
  {\includegraphics[width=\linewidth]{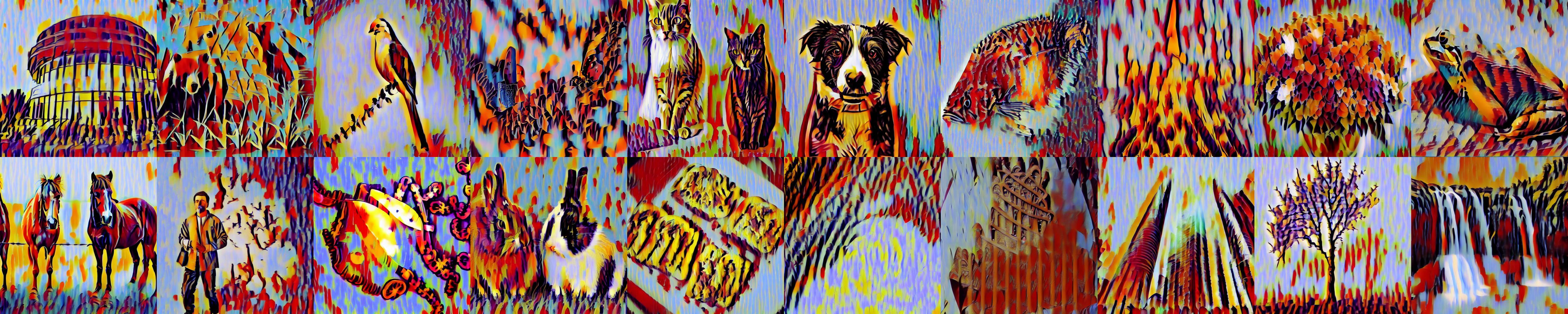}\label{fig:pre_trained_Abstractionism}}    
  \caption{Van Gogh-style and Abstractionism-style images generated by the pre-trained CDM.}
  \label{fig:canvas_images}
\end{figure}

\textbf{MU baselines:} We adopt all finetuning-based MU methods included in the UnlearnCanvas benchmark \cite{zhang2024unlearncanvas} as baselines, including CA \cite{kumari2023ablating}, Ediff \cite{wu2024erasediff}, ESD \cite{gandikota2023erasing}, Salun \cite{fan2023salun}, and SHS \cite{wu2024scissorhands}. The loss functions of Ediff, SHS, CA, ESD, and Salun, which apply mechanism \textbf{Eq.}\eqref{eq:unlearn_loss_2}, replace the unlearning images with alternative images. In this section, the unlearning art style is the Van Gogh style. For Ediff, SHS, Salun, and the proposed DiMUM, we select the Abstractionism style as the alternative art style to replace the Van Gogh style. On the other hand, the replacement styles of CA and ESD are designed manually, and we apply the default setting of \cite{zhang2024unlearncanvas}. 
Figure \ref{fig:canvas_images} demonstrates synthetic Van Gogh-style and Abstractionism-style images of the 20 classes in the UnlearnCanvas dataset, generated by the pre-trained CDM.

\begin{figure*} 
  \centering
  \subfloat[CDM unlearned by CA for 1K steps]
  {\includegraphics[width=0.48\linewidth]{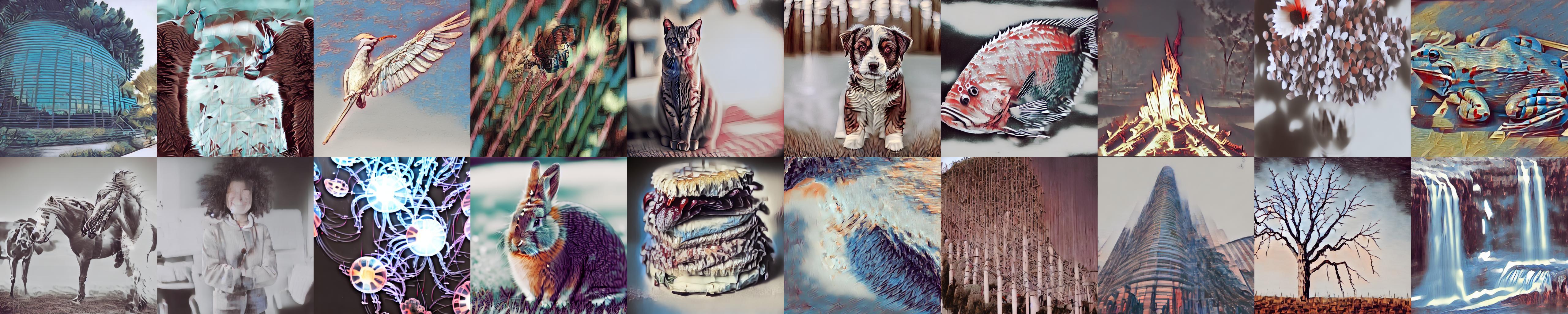}} \hspace{0.05cm}   
  \subfloat[Attacked CDM (unlearned by CA)]
  {\includegraphics[width=0.48\linewidth]{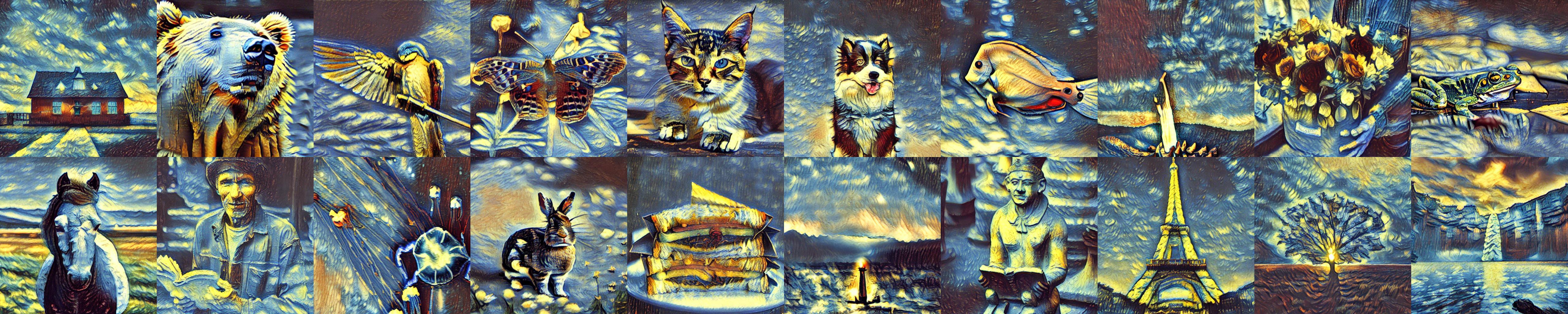}} \\ 

  \subfloat[CDM unlearned by ESD for 1K steps]
  {\includegraphics[width=0.48\linewidth]{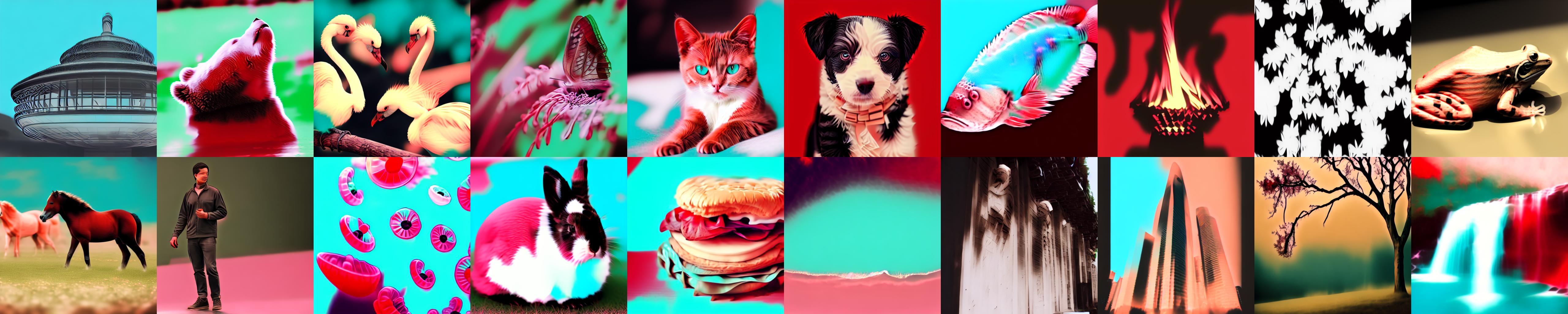}} \hspace{0.05cm}  
  \subfloat[Attacked CDM (unlearned by ESD)]
  {\includegraphics[width=0.48\linewidth]{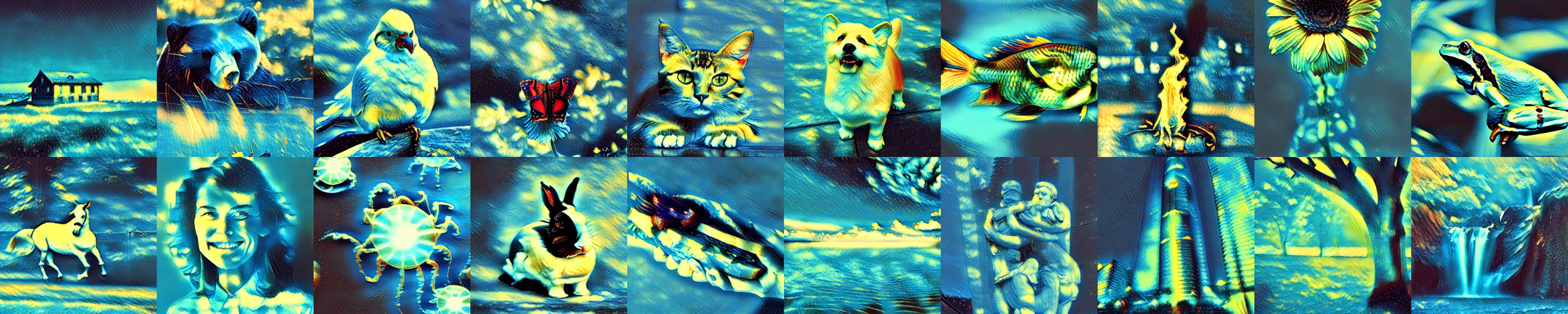}} \\ 

  \subfloat[CDM unlearned by Ediff for 1K steps]
  {\includegraphics[width=0.48\linewidth]{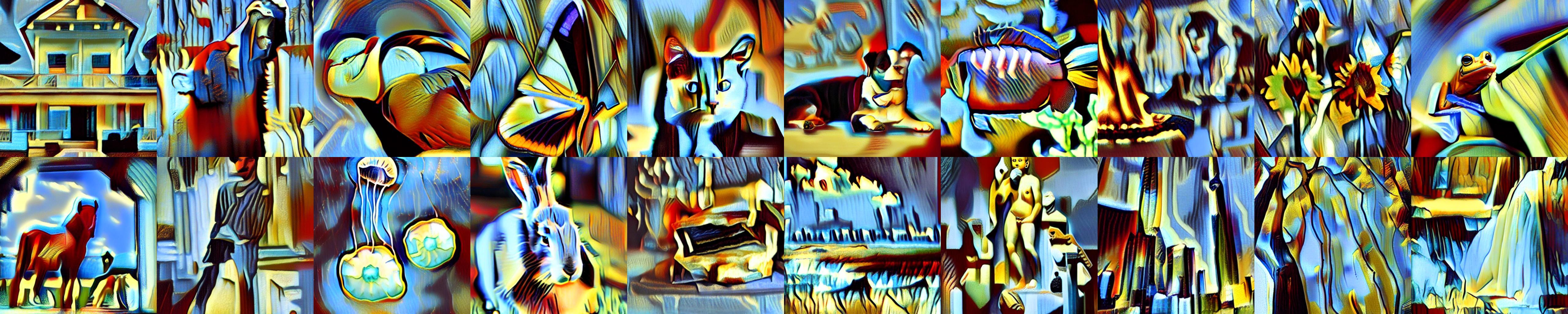}} \hspace{0.05cm}  
  \subfloat[Attacked CDM (unlearned by Ediff)]
  {\includegraphics[width=0.48\linewidth]{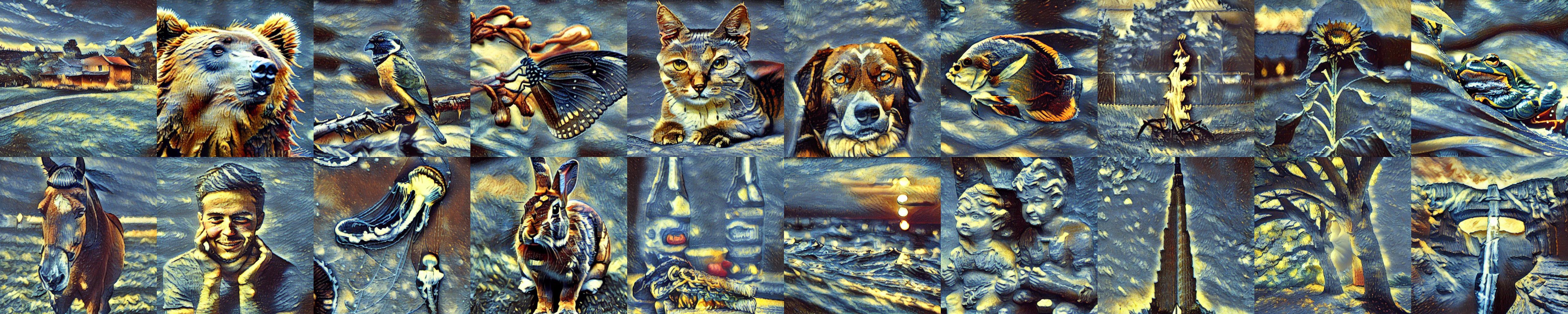}} \\ 

  \subfloat[CDM unlearned by Salun for 1K steps]
  {\includegraphics[width=0.48\linewidth]{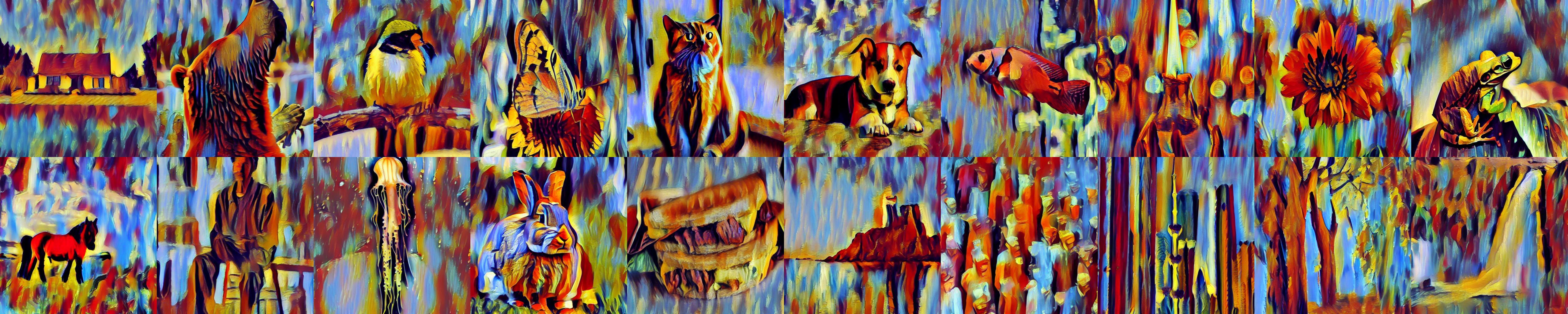}} \hspace{0.05cm}  
  \subfloat[Attacked CDM (unlearned by Salun)]
  {\includegraphics[width=0.48\linewidth]{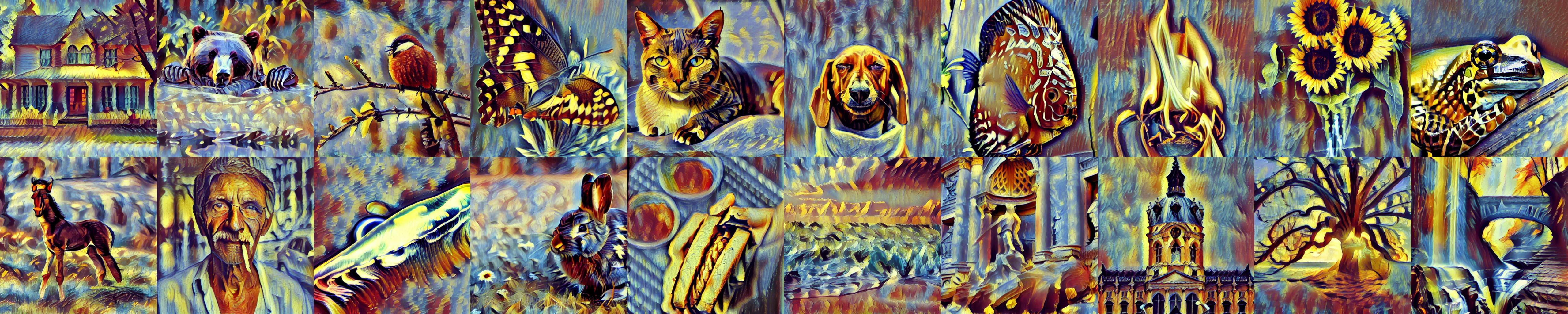}} \\ 

  \subfloat[CDM unlearned by SHS for 1K steps]
  {\includegraphics[width=0.48\linewidth]{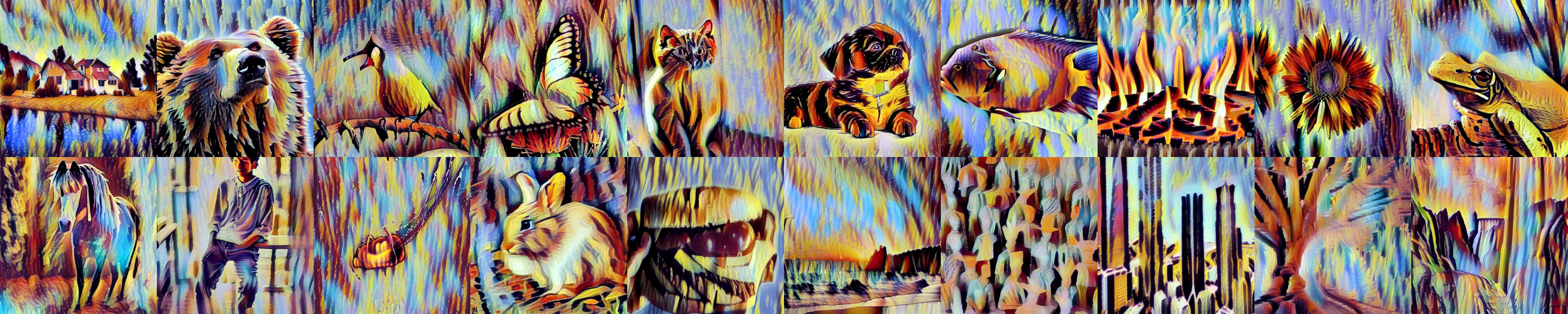}} \hspace{0.05cm}  
  \subfloat[Attacked CDM (unlearned by SHS)]
  {\includegraphics[width=0.48\linewidth]{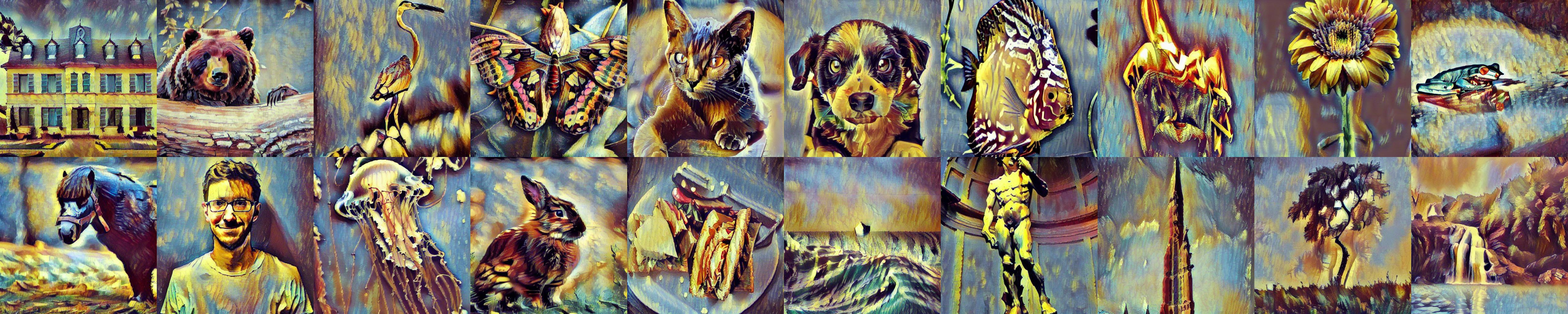}} \\ 

  \subfloat[CDM unlearned by DiMUM for 1K steps]
  {\includegraphics[width=0.48\linewidth]{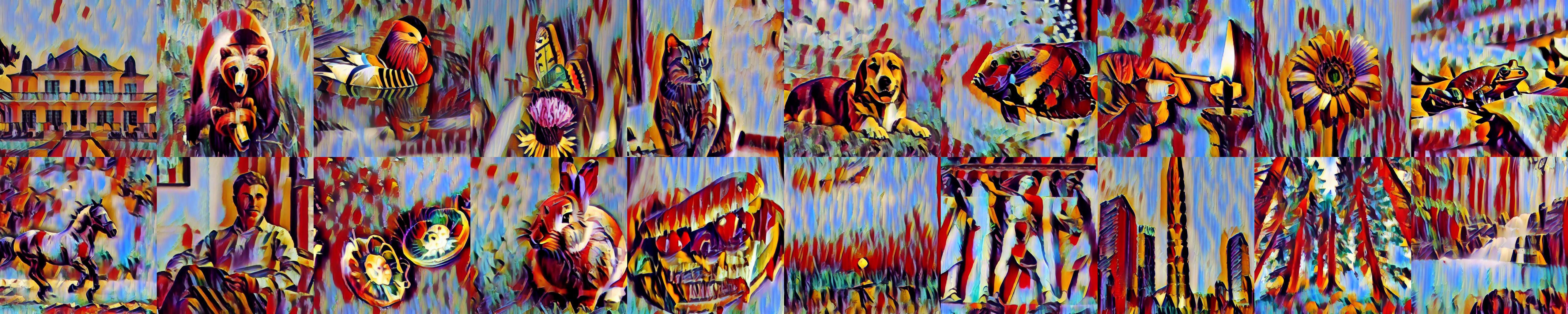} \label{fig:dimum_1k_unlearn}} \hspace{0.05cm}  
  \subfloat[Attacked CDM (unlearned by DiMUM)]
  {\includegraphics[width=0.48\linewidth]{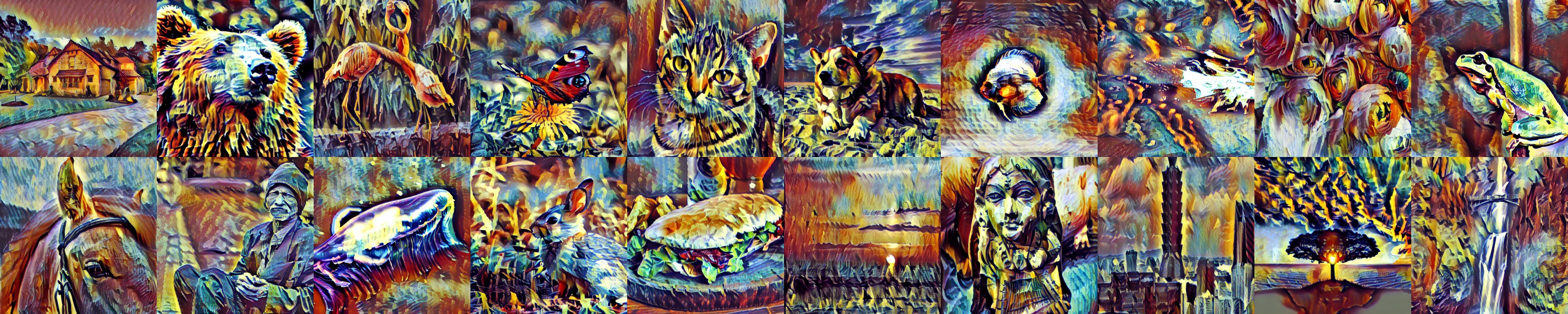}} \\ 
  \caption{Synthetic images generated by the unlearned CDMs (left side) and corresponding attacked CDMs (right side). Text prompts for all generations are "\textit{An image of \{CLASS\} in Van Gogh style.}", where Van Gogh is the unlearning art style.}
  \label{fig:canvas_images_attack}
  \vspace{-4mm}
\end{figure*}

\textbf{Evaluation process:} We directly apply the released CDM from  \cite{zhang2024unlearncanvas} as the pre-trained model $\boldsymbol\theta_p$. We construct the unlearning set $D_u$ using Van Gogh-style images from the UnlearnCanvas dataset, while the retain set $D_r$ consists of the remaining images in the UnlearnCanvas dataset.
We then input $D_u$, $D_r$, and $\boldsymbol\theta_p$ into a MU method and obtain an unlearned CDM $\boldsymbol\theta_u$. 
Next, we construct the auxiliary dataset $D_{au}$, required by DiMRA, with synthetic images generated by the pre-trained CDM, using the text prompts from the retain set $D_r$. The size of the generated $D_{au}$ is the same as that of $D_r$.
Here, we assume an attacker with limited background knowledge, as described in Assumption (ii) in Section \ref{sec:Attacker Model} and that the unlearned CDM preserves the generative capabilities of the pre-trained CDM. 
Due to the strong generative capability of the large-scale CDM, DiMRA can successfully attack the unlearned CDM with this realistic assumption.

\begin{table}
\centering
\caption{Evaluation Metrics for Different Methods and Unlearning Steps when Unlearning Van Gogh style}
\begin{tabular}{cccccc}
\hline
\textbf{Steps} & \textbf{Method} & \textbf{FID} $\downarrow$ & \textbf{AR$_{\text{MU}}$} $\downarrow$& \textbf{AR$_{\text{DiMRA}}$} $\downarrow$& \textbf{AR$_\text{CL}$ $\uparrow$}  \\
\hline
\multirow{6}*{\textbf{1K}} & {CA\cite{kumari2023ablating}} & 52.39 &0  & 1 &{{NA}}\\
{} & {ESD\cite{gandikota2023erasing}}  & 69.32 & 0 & 1 &{{NA}}\\
{} & {Ediff\cite{wu2024erasediff}}  & 71.27 &0.02  &1 & 0 \\
{} & {Salun\cite{fan2023salun}}  & 43.08  &0  & 0.78 & 0.09\\
{} & {SHS\cite{wu2024scissorhands}}  & 48.99 &0.01  & 0.60  & 0.02\\
{} & \textcolor{blue}{DiMUM(Ours)} & \textbf{42.71} &0 & \textbf{0.17} &\textbf{0.39} \\
\midrule
\multirow{6}*{\textbf{2K}} & {CA\cite{kumari2023ablating}} & 56.82 &0  & 1 &{{NA}}\\
{} & {ESD\cite{gandikota2023erasing}}   & 67.38 &0  & 0.98 &{{NA}}\\
{} & {Ediff\cite{wu2024erasediff}}   & 62.69 &0.01  & 1  & 0\\
{} & {Salun\cite{fan2023salun}}   & 41.05  &0  & 0.39  &0.09\\
{} & {SHS\cite{wu2024scissorhands}}  & 45.72 &0  & 0.35  &0.14\\
{} & \textcolor{blue}{DiMUM(Ours)}  & \textbf{40.93} &0 & \textbf{0.02} &\textbf{0.59} \\
\midrule
\multirow{6}*{\textbf{5K}} & 
{CA\cite{kumari2023ablating}} & 70.45 &0  & 0.99 &{{NA}}\\
{} & {ESD\cite{gandikota2023erasing}}  & 70.84 &0 & 0.91 &{{NA}}\\
{} & {Ediff\cite{wu2024erasediff}}  & 54.48  &0  & 0.98 & 0 \\
{} & {Salun\cite{fan2023salun}}  & \textbf{41.00}  &0  & 0.09 &0.40\\
{} & {SHS\cite{wu2024scissorhands}}  & 44.16 &0  & 0.12 &0.35\\
{} & \textcolor{blue}{DiMUM(Ours)}  & 48.07  &0 & \textbf{0.01} &\textbf{0.91} \\
\hline
\end{tabular}
\begin{tablenotes}
    \item[a] {FID and AR$_\text{MU}$ are calculated with the CDM after the last unlearning step. AR$_\text{DiMUM}$ and AR$_\text{CL}$ are the largest and smallest values, respectively, during 10K attack steps.}
\end{tablenotes}
\label{tab:matrics_van_gogh}
\end{table}

\textbf{Evaluation Metrics:}
Similar to the object unlearning case in Section \ref{sec:Unlearning a Class}, we evaluate MU methods with respect to objective (i) in Definition \ref{def:obj} using the FID between the synthetic dataset generated by the unlearned CDM and the retained set. We apply $\text{AR}_\text{MU}$ to evaluate MU methods with respect to objective (ii) in Definition \ref{def:obj}. $\text{AR}_\text{DiMRA}$ is computed to evaluate the effectiveness of the proposed DiMRA and robustness of MU methods against DiMRA. 
Here, we use the pre-trained art style classification model released in \cite{zhang2024unlearncanvas} to classify the generated images when calculating $\text{AR}_\text{MU}$ and $\text{AR}_\text{DiMRA}$.
\textbf{Note} that the image is considered correctly classified if the ground-truth label appears in the top-3 predictions of the classification model when calculating $\text{AR}_\text{MU}$ and $\text{AR}_\text{DiMRA}$, following \cite{zhang2024generate}.
In addition to the above metrics, we also evaluate the convergence of the unlearned CDMs. Specifically, we use the attacked CDMs to generate 100 images based on text prompts from the unlearning set and classify them using a pre-trained art style classification model. The accuracy rate is then computed as the proportion of the 100 generated images classified as the alternative art style. This accuracy rate is denoted as AR$_\text{CL}$.
The rational behind AR$_\text{CL}$ is that the unlearned CDMs are expected to generate images in the alternative art style given the text prompts of the unlearning style. Therefore, if the CDM is well converged, it will continue to generate images in the alternative art style given the text prompts of the unlearning style after being attacked with DiMRA.

\begin{figure*} 
  \centering
  \subfloat[CDMs unlearned by 1K steps]
  {\includegraphics[width=0.33\linewidth]{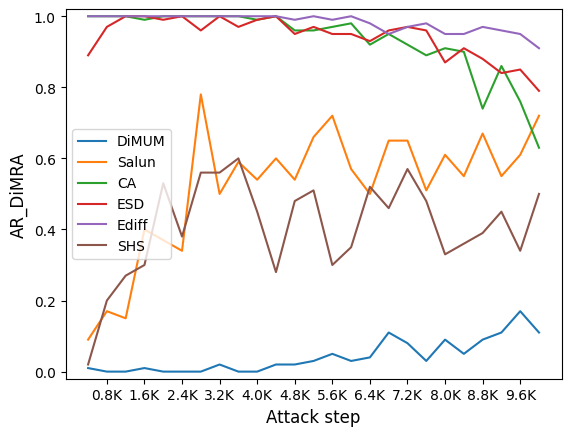}\label{fig:ar_1k}}     
  \subfloat[CDMs unlearned by 2K steps]
  {\includegraphics[width=0.33\linewidth]{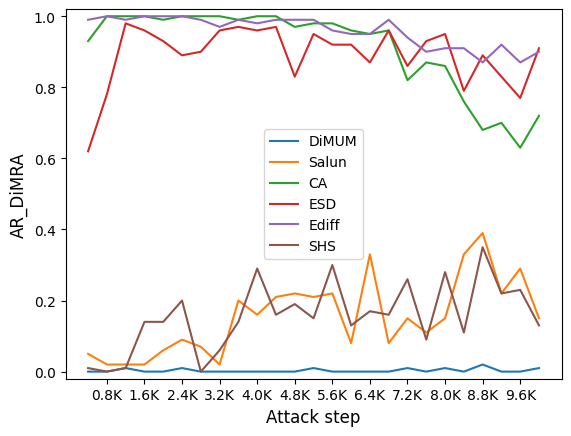}\label{fig:ar_2k}}
  \subfloat[CDMs unlearned by 5K steps]
  {\includegraphics[width=0.33\linewidth]{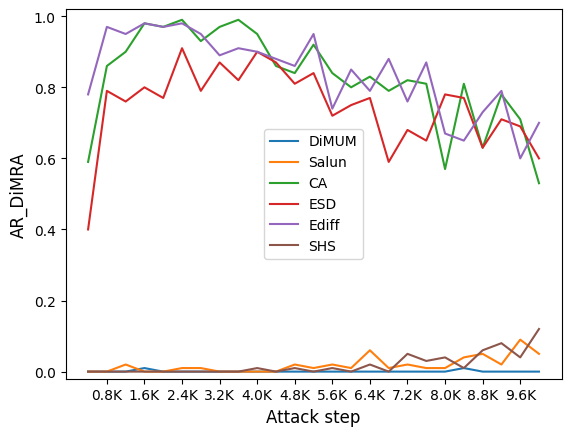}\label{fig:ar_5k}} 

  \caption{This figure shows the AR$_{\text{DiMRA}}$ curves during the attack process of DiMRA}
  \label{fig:ar_canvas}
  \vspace{-4mm}
\end{figure*}

\begin{figure*} 
  \centering
  \subfloat[CDMs unlearned by 1K steps]
  {\includegraphics[width=0.33\linewidth]{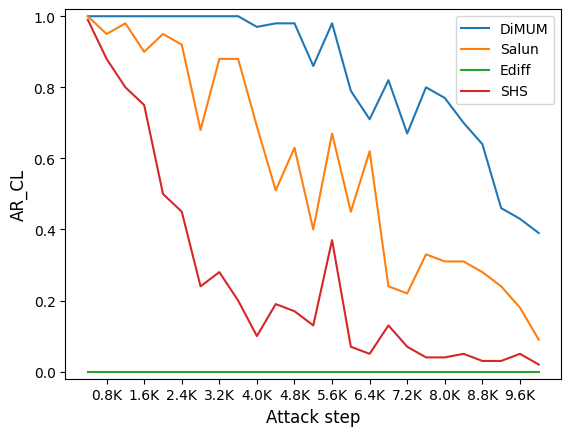}\label{fig:ar_cl_1k}}     
  \subfloat[CDMs unlearned by 2K steps]
  {\includegraphics[width=0.33\linewidth]{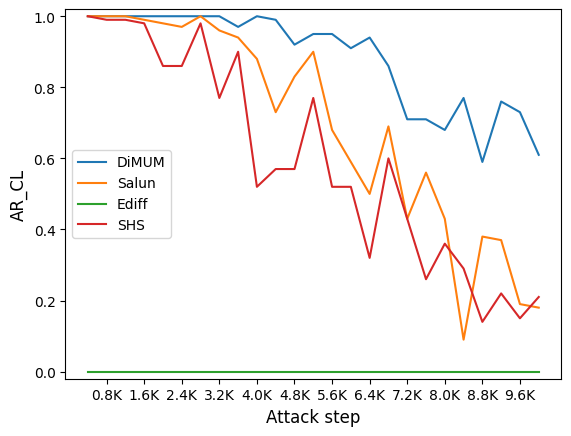}\label{fig:ar_cl_2k}}
  \subfloat[CDMs unlearned by 5K steps]
  {\includegraphics[width=0.33\linewidth]{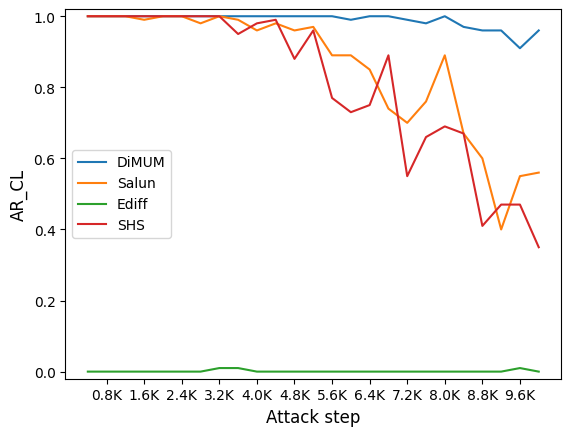}\label{fig:ar_cl_5k}}  

  \caption{This figure shows the AR$_{\text{CL}}$ curves during the attack process of DiMRA.}
  \label{fig:ar_cl_canvas}
  
\end{figure*}

\subsubsection{Evaluation of Objective (i) in Definition \ref{def:obj}}
Table \ref{tab:matrics_van_gogh} presents the FID between the synthetic dataset generated by the unlearned CDMs and the retained set. For the CDMs unlearned by CA, ESD, and Ediff, the FID scores are notably higher than those of the CDMs unlearned by other methods, indicating their unsatisfactory performance with respect to objective (i) in Definition \ref{def:obj}. The CDM unlearned by DiMUM presents the lowest FID after 1K and 2K unlearning steps. However, after 5K unlearning steps, the CDM unlearned by DiMUM exhibits a larger FID compared to those unlearned by Salun and SHS. Due to the large scale of the base model, 5K optimization steps are insufficient for the model to converge to a new local optimum with respect to both unlearning and retained losses in DiMUM. As a result, before the CDM being unlearned by DiMUM reaches a new local optimum with respect to the retain loss, its generative performance may fall below that of CDMs being unlearned by SHS and Salun.

\textbf{Takeaways:} It is a common assumption that the unlearning step should be significantly shorter than the pre-training step; otherwise, MU offers no advantage over retraining the model from scratch. In this experiment, 1K and 2K unlearning steps are appropriate, as the pre-trained model is fine-tuned on the UnlearnCanvas dataset for only 20K steps \cite{zhang2024unlearncanvas}. Additionally, we conduct experiments with MU methods for 5K steps to assess their performance under a larger unlearning step.

\subsubsection{Evaluation of Objective (ii) in Definition \ref{def:obj}}
As shown by AR$_\text{MU}$ in Table \ref{tab:matrics_van_gogh}, all MU methods successfully prevent the generation of Van Gogh-style images, even with just 1K unlearning steps. However, AR$_\text{DiMRA}$ values in Table \ref{tab:matrics_van_gogh} indicate that CA, ESD, and Ediff are highly vulnerable to the proposed DiMRA, as over 90\% synthetic images generated by the attacked CDM are classified as Van Gogh style even though the CDM has been unlearned for 5K steps. On the other hand, the proposed DiMUM is the most robust against DiMRA, with only 2\% of synthetic images generated by the attacked CDM, which has been unlearned by DiMUM for 2K steps, classified as Van Gogh style. Figure \ref{fig:canvas_images_attack} shows the synthetic images generated by the unlearned CDMs (left side) and the corresponding attacked CDMs (right side), where the number of unlearning steps is 1K and the attack step is 10K. The results are consistent with those in Table \ref{tab:matrics_van_gogh}, where DiMUM is the most resistant to DiMRA, while CA, ESD, and Ediff are highly vulnerable. Figure \ref{fig:ar_canvas} shows the AR$_\text{DiMRA}$ curves during 10K attack steps. As illustrated, 400 attack steps can effectively reverse the CA, ESD, and Ediff. In contrast, DiMUM presents significantly lower AR$_\text{DiMRA}$ than other baselines.

Next, we evaluate the convergence of CDMs unlearned by Ediff, Salun, SHS, and DiMUM. The alternative images of CA and ESD are manually designed and thus cannot be classified by the pre-trained art-style classification model. Thus, they are not included in this experiment. 
Comparing the left side of Fig. \ref{fig:canvas_images_attack} with Fig. \ref{fig:pre_trained_Abstractionism}, the CDM unlearned by DiMUM generates the Abstractionism-style images most similar to those produced by the pre-trained model, given text prompts \textit{"An image of \{CLASS\} in Van Gogh style."}.
Thus, the CDM unlearned by DiMUM converges most to the unlearning objective, i.e., replacing the unlearning art style with an alternative art style, compared to those unlearned by Ediff, Salun, and SHS.
Figure \ref{fig:ar_cl_canvas} shows the AR$_\text{CL}$ curves during 10K attack steps. As shown, CDMs unlearned by Ediff, Salun, and SHS are more easily reversed, i.e., they forget the alternative art style during the attack, indicating poor convergence. In contrast, over 90\% of the synthetic images generated by the CDM, which has been unlearned by DiMUM for 5K steps, are classified as Abstractionism style during the 10K attack steps given text prompts \textit{"An image of \{CLASS\} in Van Gogh style."}. This also indicates that the CDM unlearned by DiMUM achieves the best convergence, as the generation distribution remains largely unchanged during the attack.

\begin{tcolorbox}[colback=blue!5!white, colframe=blue!75!black, title=Takeaways]
Comparing the AR$_\text{DiMRA}$ curves in Fig. \ref{fig:ar_canvas} with the AR$_\text{CL}$ curves in Fig. \ref{fig:ar_cl_canvas}, we observe that a higher AR$_\text{CL}$ (convergence to the unlearning loss) typically corresponds to a lower AR$_\text{DiMRA}$, suggesting that the convergent unlearning loss of DiMUM enhances robustness against DiMRA. Therefore, we recommend that future research about MU for CDMs prioritize the convergence of the unlearning loss, rather than merely degrading the generative ability of CDMs for the unlearning concepts.
\end{tcolorbox}


\section{Conclusion}
In this paper, we first introduce a novel relearning attack for machine unlearning (MU) in conditional diffusion models (CDMs), called the Diffusion Model Relearning Attack (DiMRA). DiMRA finetunes the unlearned models on an auxiliary dataset that can be easily constructed if the CDM is publicly available. Experimental results show that the attacked CDMs regenerate previously unlearned elements, highlighting the vulnerability of current MU techniques. Second, we propose a novel MU method for CDMs, Diffusion Model Unlearning by Memorization (DiMUM), which unlearns targeted elements by memorizing alternative elements given conditioning inputs from the unlearning set. Unlike other MU methods, DiMUM achieves convergence with respect to both unlearning and retain losses. Experimental results demonstrate that DiMUM can optimize the CDM's parameters to a new local optimum and is the most robust against DiMRA.



\bibliographystyle{IEEEtran}
\bibliography{reference}

@misc{EU_GDPR_2016,
  title        = {Regulation (EU) 2016/679 of the European Parliament and of the Council of 27 April 2016 on the protection of natural persons with regard to the processing of personal data and on the free movement of such data (General Data Protection Regulation)},
  howpublished = {Official Journal of the European Union (OJ L 119, 4.5.2016, pp. 1–88)},
  year         = {2016},
  month        = {Apr.},
  note         = {Accessed: Oct. 24, 2025. [Online]. Available: \url{https://eur-lex.europa.eu/eli/reg/2016/679/oj}}
}

@inproceedings{rombach2022high,
  title={High-resolution image synthesis with latent diffusion models},
  author={Rombach, Robin and Blattmann, Andreas and Lorenz, Dominik and Esser, Patrick and Ommer, Bj{\"o}rn},
  booktitle={Proceedings of the IEEE/CVF conference on computer vision and pattern recognition},
  pages={10684--10695},
  year={2022}
}

@article{saharia2022photorealistic,
  title={Photorealistic text-to-image diffusion models with deep language understanding},
  author={Saharia, Chitwan and Chan, William and Saxena, Saurabh and Li, Lala and Whang, Jay and Denton, Emily L and Ghasemipour, Kamyar and Gontijo Lopes, Raphael and Karagol Ayan, Burcu and Salimans, Tim and others},
  journal={Advances in neural information processing systems},
  volume={35},
  pages={36479--36494},
  year={2022}
}

@inproceedings{lugmayr2022repaint,
  title={Repaint: Inpainting using denoising diffusion probabilistic models},
  author={Lugmayr, Andreas and Danelljan, Martin and Romero, Andres and Yu, Fisher and Timofte, Radu and Van Gool, Luc},
  booktitle={Proceedings of the IEEE/CVF conference on computer vision and pattern recognition},
  pages={11461--11471},
  year={2022}
}

@inproceedings{saharia2022palette,
  title={Palette: Image-to-image diffusion models},
  author={Saharia, Chitwan and Chan, William and Chang, Huiwen and Lee, Chris and Ho, Jonathan and Salimans, Tim and Fleet, David and Norouzi, Mohammad},
  booktitle={ACM SIGGRAPH 2022 conference proceedings},
  pages={1--10},
  year={2022}
}

@article{kazerouni2023diffusion,
  title={Diffusion models in medical imaging: A comprehensive survey},
  author={Kazerouni, Amirhossein and Aghdam, Ehsan Khodapanah and Heidari, Moein and Azad, Reza and Fayyaz, Mohsen and Hacihaliloglu, Ilker and Merhof, Dorit},
  journal={Medical Image Analysis},
  volume={88},
  pages={102846},
  year={2023},
  publisher={Elsevier}
}

@inproceedings{wolleb2022diffusion,
  title={Diffusion models for medical anomaly detection},
  author={Wolleb, Julia and Bieder, Florentin and Sandk{\"u}hler, Robin and Cattin, Philippe C},
  booktitle={International Conference on Medical image computing and computer-assisted intervention},
  pages={35--45},
  year={2022},
  organization={Springer}
}

@article{ho2020denoising,
  title={Denoising diffusion probabilistic models},
  author={Ho, Jonathan and Jain, Ajay and Abbeel, Pieter},
  journal={Advances in neural information processing systems},
  volume={33},
  pages={6840--6851},
  year={2020}
}

@inproceedings{nichol2021improved,
  title={Improved denoising diffusion probabilistic models},
  author={Nichol, Alexander Quinn and Dhariwal, Prafulla},
  booktitle={International conference on machine learning},
  pages={8162--8171},
  year={2021},
  organization={PMLR}
}

@inproceedings{schramowski2023safe,
  title={Safe latent diffusion: Mitigating inappropriate degeneration in diffusion models},
  author={Schramowski, Patrick and Brack, Manuel and Deiseroth, Bj{\"o}rn and Kersting, Kristian},
  booktitle={Proceedings of the IEEE/CVF Conference on Computer Vision and Pattern Recognition},
  pages={22522--22531},
  year={2023}
}

@article{mittica2023ai,
  title={AI and Artworks: Legal and Technical Issues},
  author={Mittica, Francesco},
  journal={Available at SSRN 4826142},
  year={2023}
}

@article{golda2024privacy,
  title={Privacy and Security Concerns in Generative AI: A Comprehensive Survey},
  author={Golda, Abenezer and Mekonen, Kidus and Pandey, Amit and Singh, Anushka and Hassija, Vikas and Chamola, Vinay and Sikdar, Biplab},
  journal={IEEE Access},
  year={2024},
  publisher={IEEE}
}

@article{huang2024unified,
  title={Unified Gradient-Based Machine Unlearning with Remain Geometry Enhancement},
  author={Huang, Zhehao and Cheng, Xinwen and Zheng, JingHao and Wang, Haoran and He, Zhengbao and Li, Tao and Huang, Xiaolin},
  journal={arXiv preprint arXiv:2409.19732},
  year={2024}
}

@article{heng2024selective,
  title={Selective amnesia: A continual learning approach to forgetting in deep generative models},
  author={Heng, Alvin and Soh, Harold},
  journal={Advances in Neural Information Processing Systems},
  volume={36},
  year={2024}
}

@article{fan2023salun,
  title={Salun: Empowering machine unlearning via gradient-based weight saliency in both image classification and generation},
  author={Fan, Chongyu and Liu, Jiancheng and Zhang, Yihua and Wong, Eric and Wei, Dennis and Liu, Sijia},
  journal={arXiv preprint arXiv:2310.12508},
  year={2023}
}

@inproceedings{gandikota2024unified,
  title={Unified concept editing in diffusion models},
  author={Gandikota, Rohit and Orgad, Hadas and Belinkov, Yonatan and Materzy{\'n}ska, Joanna and Bau, David},
  booktitle={Proceedings of the IEEE/CVF Winter Conference on Applications of Computer Vision},
  pages={5111--5120},
  year={2024}
}

@inproceedings{kumari2023ablating,
  title={Ablating concepts in text-to-image diffusion models},
  author={Kumari, Nupur and Zhang, Bingliang and Wang, Sheng-Yu and Shechtman, Eli and Zhang, Richard and Zhu, Jun-Yan},
  booktitle={Proceedings of the IEEE/CVF International Conference on Computer Vision},
  pages={22691--22702},
  year={2023}
}

@article{wu2024erasediff,
  title={Erasediff: Erasing data influence in diffusion models},
  author={Wu, Jing and Le, Trung and Hayat, Munawar and Harandi, Mehrtash},
  journal={arXiv preprint arXiv:2401.05779},
  year={2024}
}

@inproceedings{gandikota2023erasing,
  title={Erasing concepts from diffusion models},
  author={Gandikota, Rohit and Materzynska, Joanna and Fiotto-Kaufman, Jaden and Bau, David},
  booktitle={Proceedings of the IEEE/CVF International Conference on Computer Vision},
  pages={2426--2436},
  year={2023}
}

@inproceedings{zhang2024forget,
  title={Forget-me-not: Learning to forget in text-to-image diffusion models},
  author={Zhang, Gong and Wang, Kai and Xu, Xingqian and Wang, Zhangyang and Shi, Humphrey},
  booktitle={Proceedings of the IEEE/CVF Conference on Computer Vision and Pattern Recognition},
  pages={1755--1764},
  year={2024}
}

@article{li2024get,
  title={Get What You Want, Not What You Don't: Image Content Suppression for Text-to-Image Diffusion Models},
  author={Li, Senmao and van de Weijer, Joost and Hu, Taihang and Khan, Fahad Shahbaz and Hou, Qibin and Wang, Yaxing and Yang, Jian},
  journal={arXiv preprint arXiv:2402.05375},
  year={2024}
}

@inproceedings{wu2024scissorhands,
  title={Scissorhands: Scrub data influence via connection sensitivity in networks},
  author={Wu, Jing and Harandi, Mehrtash},
  booktitle={European Conference on Computer Vision},
  pages={367--384},
  year={2024},
  organization={Springer}
}

@inproceedings{lyu2024one,
  title={One-dimensional Adapter to Rule Them All: Concepts Diffusion Models and Erasing Applications},
  author={Lyu, Mengyao and Yang, Yuhong and Hong, Haiwen and Chen, Hui and Jin, Xuan and He, Yuan and Xue, Hui and Han, Jungong and Ding, Guiguang},
  booktitle={Proceedings of the IEEE/CVF Conference on Computer Vision and Pattern Recognition},
  pages={7559--7568},
  year={2024}
}

@inproceedings{li2024safegen,
  title={Safegen: Mitigating sexually explicit content generation in text-to-image models},
  author={Li, Xinfeng and Yang, Yuchen and Deng, Jiangyi and Yan, Chen and Chen, Yanjiao and Ji, Xiaoyu and Xu, Wenyuan},
  booktitle={Proceedings of the 2024 on ACM SIGSAC Conference on Computer and Communications Security},
  pages={4807--4821},
  year={2024}
}

@article{zhang2024unlearncanvas,
  title={Unlearncanvas: A stylized image dataset to benchmark machine unlearning for diffusion models},
  author={Zhang, Yihua and Zhang, Yimeng and Yao, Yuguang and Jia, Jinghan and Liu, Jiancheng and Liu, Xiaoming and Liu, Sijia},
  journal={arXiv preprint arXiv:2402.11846},
  year={2024}
}

@inproceedings{
song2020denoising,
title={Denoising Diffusion Implicit Models},
author={Jiaming Song and Chenlin Meng and Stefano Ermon},
booktitle={International Conference on Learning Representations},
year={2021},
url={https://openreview.net/forum?id=St1giarCHLP}
}

@inproceedings{ho2022classifier,
  title={Classifier-Free Diffusion Guidance},
  author={Ho, Jonathan and Salimans, Tim},
  booktitle={NeurIPS 2021 Workshop on Deep Generative Models and Downstream Applications},
  year={2021}
}

@article{baptista2025memorization,
  title={Memorization and Regularization in Generative Diffusion Models},
  author={Baptista, Ricardo and Dasgupta, Agnimitra and Kovachki, Nikola B and Oberai, Assad and Stuart, Andrew M},
  journal={arXiv preprint arXiv:2501.15785},
  year={2025}
}

@inproceedings{yoon2023diffusion,
  title={Diffusion probabilistic models generalize when they fail to memorize},
  author={Yoon, TaeHo and Choi, Joo Young and Kwon, Sehyun and Ryu, Ernest K},
  booktitle={ICML 2023 workshop on structured probabilistic inference $\{$$\backslash$\&$\}$ generative modeling},
  year={2023}
}

@article{gu2023memorization,
  title={On memorization in diffusion models},
  author={Gu, Xiangming and Du, Chao and Pang, Tianyu and Li, Chongxuan and Lin, Min and Wang, Ye},
  journal={arXiv preprint arXiv:2310.02664},
  year={2023}
}

@inproceedings{zhang2024emergence,
  title={The emergence of reproducibility and consistency in diffusion models},
  author={Zhang, Huijie and Zhou, Jinfan and Lu, Yifu and Guo, Minzhe and Wang, Peng and Shen, Liyue and Qu, Qing},
  booktitle={Forty-first International Conference on Machine Learning},
  year={2024}
}

@article{heusel2017gans,
  title={Gans trained by a two time-scale update rule converge to a local nash equilibrium},
  author={Heusel, Martin and Ramsauer, Hubert and Unterthiner, Thomas and Nessler, Bernhard and Hochreiter, Sepp},
  journal={Advances in neural information processing systems},
  volume={30},
  year={2017}
}

@inproceedings{
dhariwal2021diffusion,
title={Diffusion Models Beat {GAN}s on Image Synthesis},
author={Prafulla Dhariwal and Alexander Quinn Nichol},
booktitle={Advances in Neural Information Processing Systems},
editor={A. Beygelzimer and Y. Dauphin and P. Liang and J. Wortman Vaughan},
year={2021},
url={https://openreview.net/forum?id=AAWuCvzaVt}
}

@inproceedings{chin2023prompting4debugging,
  author = {Chin, Zhi-Yi and Jiang, Chieh-Ming and Huang, Ching-Chun and Chen, Pin-Yu and Chiu, Wei-Cheng},
title = {Prompting4Debugging: red-teaming text-to-image diffusion models by finding problematic prompts},
year = {2024},
publisher = {JMLR.org},
articleno = {336},
numpages = {19},
location = {Vienna, Austria},
series = {ICML'24}
}

@misc{maus2023blackboxadversarialprompting,
      title={Black Box Adversarial Prompting for Foundation Models}, 
      author={Natalie Maus and Patrick Chao and Eric Wong and Jacob Gardner},
      year={2023},
      eprint={2302.04237},
      archivePrefix={arXiv},
      primaryClass={cs.LG},
      url={https://arxiv.org/abs/2302.04237}, 
}

@inproceedings{zhang2024generate,
  title={To generate or not? safety-driven unlearned diffusion models are still easy to generate unsafe images... for now},
  author={Zhang, Yimeng and Jia, Jinghan and Chen, Xin and Chen, Aochuan and Zhang, Yihua and Liu, Jiancheng and Ding, Ke and Liu, Sijia},
  booktitle={European Conference on Computer Vision},
  pages={385--403},
  year={2024},
  organization={Springer}
}

@article{yao2024large,
  title={Large language model unlearning},
  author={Yao, Yuanshun and Xu, Xiaojun and Liu, Yang},
  journal={Advances in Neural Information Processing Systems},
  volume={37},
  pages={105425--105475},
  year={2024}
}

@inproceedings{thudi2022unrolling,
  title={Unrolling sgd: Understanding factors influencing machine unlearning},
  author={Thudi, Anvith and Deza, Gabriel and Chandrasekaran, Varun and Papernot, Nicolas},
  booktitle={2022 IEEE 7th European Symposium on Security and Privacy (EuroS\&P)},
  pages={303--319},
  year={2022},
  organization={IEEE}
}

@inproceedings{
cha2025towards,
title={Towards Robust and Parameter-Efficient Knowledge Unlearning for {LLM}s},
author={Sungmin Cha and Sungjun Cho and Dasol Hwang and Moontae Lee},
booktitle={The Thirteenth International Conference on Learning Representations},
year={2025},
url={https://openreview.net/forum?id=1ExfUpmIW4}
}

@ARTICLE{10113700,
  author={Tarun, Ayush K. and Chundawat, Vikram S. and Mandal, Murari and Kankanhalli, Mohan},
  journal={IEEE Transactions on Neural Networks and Learning Systems}, 
  title={Fast Yet Effective Machine Unlearning}, 
  year={2024},
  volume={35},
  number={9},
  pages={13046-13055},
  doi={10.1109/TNNLS.2023.3266233}}

\end{document}